\documentclass[11pt]{article}
\PassOptionsToPackage{hyphens}{url}
\usepackage{acl}
\usepackage{times}
\usepackage{latexsym}
\usepackage[T1]{fontenc}
\usepackage[utf8]{inputenc}
\usepackage{microtype}
\usepackage{inconsolata}

\usepackage{graphicx}
\newsavebox{\supporttablebox}
\newenvironment{supporttable}{%
  \begin{lrbox}{\supporttablebox}\small\ignorespaces
}{%
  \unskip\end{lrbox}%
  \ifdim\wd\supporttablebox>\textwidth
    \resizebox{\textwidth}{!}{\usebox{\supporttablebox}}%
  \else
    \usebox{\supporttablebox}%
  \fi
}
\usepackage{amsmath,amssymb,amsthm}
\usepackage{booktabs,multirow,pifont}
\usepackage{tablefootnote,makecell}
\usepackage{enumitem}
\usepackage{epigraph,float,colortbl}
\usepackage[most]{tcolorbox}
\usepackage{tikz}
\definecolor{gray94}{gray}{.94}
\definecolor{gray90}{gray}{.90}
\definecolor{basegray}{RGB}{245,245,245}
\definecolor{rlvrblue}{RGB}{230,245,255}

\definecolor{lightred}{RGB}{255,240,240}
\definecolor{lightgreen}{RGB}{240,255,240}
\definecolor{lightblue}{RGB}{240,245,255}
\definecolor{tablelightgray}{RGB}{245,245,245}
\definecolor{MyGreen}{RGB}{34,139,34}
\definecolor{myblue}{RGB}{0,77,64} % dark greenish
\definecolor{mybg}{RGB}{240,248,247} % light bluish background

\newtcolorbox{takeawaybox}{
  enhanced,
  colback=mybg,
  colframe=myblue,
  coltitle=myblue,
  boxrule=0.6pt,
  arc=6pt,
  left=6pt,
  right=6pt,
  top=6pt,
  bottom=6pt,
  title=\textsc{Takeaway}
}
\theoremstyle{plain}
\newtheorem{theorem}{Theorem}[section]
\newtheorem{proposition}[theorem]{Proposition}

\newtheorem{corollary}[theorem]{Corollary}
\theoremstyle{definition}
\newtheorem{definition}{Definition}

\theoremstyle{remark}

\title{The Invisible Leash? Why RLVR May or May Not Escape Its Origin}
\author{
  \textbf{Fang Wu}\textsuperscript{1}\thanks{Equal contribution.} \quad
  \textbf{Weihao Xuan}\textsuperscript{2,3}\footnotemark[1] \quad
  \textbf{Danny Xing}\textsuperscript{1} \quad
  \textbf{Ximing Lu}\textsuperscript{4,5} \quad
  \textbf{Mingjie Liu}\textsuperscript{5} \\
  \textbf{Yi Dong}\textsuperscript{5} \quad
  \textbf{Zaid Harchaoui}\textsuperscript{4} \quad
  \textbf{Yejin Choi}\textsuperscript{1} \\[4pt]
  \textsuperscript{1}Stanford University, USA \quad
  \textsuperscript{2}University of Tokyo, Japan \quad
  \textsuperscript{3}RIKEN AIP, Japan \\
  \textsuperscript{4}University of Washington, USA \quad
  \textsuperscript{5}NVIDIA, USA \\[4pt]
  \small
  \texttt{\{fangwu97,yejinc\}@stanford.edu},
  \texttt{xuan@ms.k.u-tokyo.ac.jp},
  \texttt{lux32@cs.washington.edu}, \\
  \small
  \texttt{zaid@uw.edu},
  \texttt{\{yidong,mingjiel\}@nvidia.com}
}

\begin{document}

\maketitle

\begin{abstract}
    Recent advances highlight Reinforcement Learning with Verifiable Rewards (RLVR) as a promising approach to enhancing LLM capabilities. However, it remains unclear whether the current practice of RLVR truly expands models' reasoning boundary or mainly amplifies high-reward outputs that base models already know. We introduce empirical support dynamics, a framework that quantifies how RLVR preserves, expands, and shrinks access to verified solutions. Applied across models, reasoning tasks, and training checkpoints, it matches verified successes question by question to reveal changes that aggregate accuracy can conceal. Across the evaluated model-level benchmark aggregates, \textit{the shrinkage of observed support outweighs its expansion}: more questions lose a verified success than gain one at the stated sampling budgets. These losses occur alongside gains in \texttt{pass@1}. Some RLVR models increase average token-level entropy while reducing final-answer entropy: their token choices become more uncertain on average even as their final answers become more concentrated. Together, these findings reveal an ``invisible leash'' in current RLVR practice: it can make existing successes more reliable while losing access to other verified solutions. Empirical support dynamics make this imbalance measurable and motivate exploration strategies that reach underrepresented solutions while retaining prior successes.
\end{abstract}


\section{Introduction}

The rise of large reasoning models marks a breakthrough in AI capabilities, particularly in solving mathematics~\citep{zeng2025simplerlzooinvestigatingtamingzero} and programming~\citep{deepcoder2025,code-r1}.
The key ingredient behind this remarkable progress is large-scale {Reinforcement Learning with Verifiable Rewards} ({RLVR}), where a pretrained base model is optimized via RL using simple, automatically computed rewards. Prior work has also explored stronger notions of verifiability, such as models that can generate interactive proofs of their own correctness~\citep{AmitGPR25}. %  such as DeepSeek-R1~\citep{guo2025deepseek} and OpenAI-o1~\citep{jaech2024openai}
Despite empirical success, a fundamental question remains under active debate within the research community: \emph{Does the current practice of RLVR expand base models’ reasoning capabilities, or simply reinforce patterns base models already know, sometimes at the expense of exploring alternative correct solutions?}

Recent studies revealed a puzzling pattern that hints at this limitation. While RLVR models consistently outperform base models on a single attempt, base models often perform better with multiple attempts.~\citet{shao2025spurious} also suggests that RLVR even benefits from noisy or spurious reward signals, further calling into question whether observed gains reflect deeper reasoning improvements or merely sharper exploitation of existing heuristics.

We use \texttt{pass@k} to measure access to verified solutions under a specified sampling budget~\citep{chen2021evaluating,shao2024deepseekmath,chen2025pass}. Aggregate performance can conceal which verified successes are preserved, gained, or lost, and how this composition changes during training.
Some studies interpret \texttt{pass@k} as evidence that RLVR primarily performs conservative optimization within the base model's capabilities~\citep{yue2025does,zhao2025echo,shah2025rethinking,ma2025learning,he2025rewarding}. Others argue that this pattern only appears in specialized domains where base models were well-trained, and RLVR can substantially expand reasoning in other domains~\citep{liu2025prorl}.

Seeking a definitive answer to this debate remains an open challenge. In the extreme case, it seems unlikely that RLVR can, out-of-the-box, unlock advanced reasoning capabilities in any model, such as GPT-2~\citep{radford2019language}. To understand if any inherent limitations in current RLVR practice, this paper provides a systematic empirical investigation into the fundamental capabilities and potential limitations of RLVR. We introduce \emph{empirical support}: the set of questions on which a model produces at least one verified success under a specified sampling budget and evaluation protocol. Using this framework, we show that:
\begin{enumerate}[left=0pt]
    \item \textbf{The current RLVR recipe primarily preserves rather than expands the base model's solution coverage.} Across the model-level benchmark aggregates, RLVR loses observed success on more questions than it gains, even while improving single-sample accuracy. Training-stage analysis shows that expansion can be temporary: AIME25 and Minerva gain coverage at intermediate checkpoints but end below their initial coverage.
    
    \item \textbf{The support pattern extends across mathematical and non-mathematical reasoning tasks.} We observe preservation-dominated support across mathematics, logical reasoning, factual QA, and code generation.
    
    \item \textbf{Local uncertainty and answer concentration can increase together.} Some RLVR models increase average token-level entropy while reducing final-answer entropy, revealing a separation between uncertainty during generation and concentration in extracted answers.
\end{enumerate}
These findings characterize the ``invisible leash'' through measurable changes in access to verified solutions. Empirical support dynamics separates expansion from shrinkage even when they offset each other in aggregate performance. For future exploration strategies, the goal is to make successful responses in low-probability regions easier to reach while retaining prior successes.

\newif\ifAAAIfigureone
\AAAIfigureonefalse
\ifAAAIfigureone
\begin{figure}[t]
    \centering
    % \vspace{-4em}
    \includegraphics[width=\columnwidth]{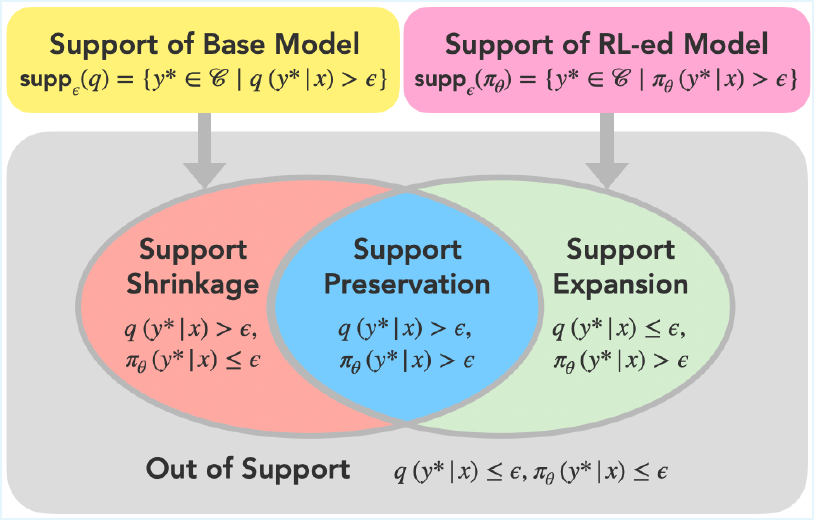}
    \vspace{-0.5em}
    \caption{Illustration of empirical support. We define four regions based on whether a correct completion $y^* \in \mathcal{C}$ is assigned non-negligible probability mass by the base and RLVR models, $q$ and $\pi_\theta$. 
    \emph{Preservation} covers completions with $q(y^*|x) > \epsilon$ and $\pi_\theta(y^*|x) > \epsilon$; 
    \emph{\textcolor{red}{Shrinkage}} includes correct completions downweighted by RLVR below $\epsilon$; 
    \emph{\textcolor{MyGreen}{Expansion}} includes completions that RLVR newly upweights above $\epsilon$ despite negligible base model mass; and \emph{\textcolor{gray}{Out of Support}} refers to completions missed by both.}
    \label{fig:fig1}
    % \vspace{-2em}
\end{figure}

\else
\begin{figure*}[!t]
    \centering
    \includegraphics[width=.93\textwidth]{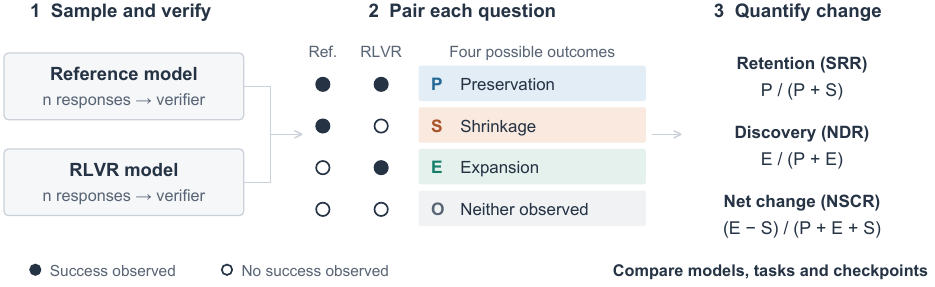}
    \vspace{-0.5em}
    \caption{Empirical support dynamics compares observed solution access question by question. Both models are evaluated on the same questions under a fixed response budget, decoding protocol, and verifier. Pairing whether each model produces at least one verified success yields preservation, shrinkage, expansion, and out of support. Their counts quantify retention, discovery, and net change across models, tasks, and training checkpoints.}
    \vspace{-0.5em}
    \label{fig:fig1}
\end{figure*}
\fi

\section{Empirical Support Dynamics}

\subsection{Formalizing Solution Accessibility}

\paragraph{Models and Verified Success.} Let $\mathcal{X}$ denote the space of natural language prompts, and $\mathcal{Y}$ denote the space of token sequences (\emph{e.g.}, reasoning traces or completions). For a fixed prompt $x \in \mathcal{X}$, $q(y \mid x)$ is the output distribution of the base model, and $R(x, y) \in \{0, 1\}$ is a verifiable reward function indicating whether $y$ is a correct solution. Various RLVR algorithms, including PPO~\citep{schulman2017proximal}, RLOO~\citep{kool2019buy}, GRPO~\citep{guo2025deepseek}, DAPO~\citep{yu2025dapo}, or REINFORCE++~\citep{hu2025reinforce}, learn a new distribution $\pi_\theta(y \mid x)$  to optimize different variants of the following regularized objective:
\[
\begin{aligned}
&\max_{\theta}\mathbb{E}_{y \sim \pi_\theta(\cdot \mid x),x \sim \mathcal{D}}
\Bigl[ R(x, y)\\
&\qquad{}-\beta^{-1} \log \frac{\pi_\theta(y \mid x)}{q(y \mid x)} \Bigr],
\end{aligned}
\] where $\mathcal{D}$ is the prompt distribution. An optional log ratio penalizes divergence from $q$ controlled by a hyperparameter $\beta > 0$. 
\paragraph{From Exact Support to Practical Access.} With finite logits, an unmasked softmax gives positive probability to every token and hence every finite completion permitted by the generation rules. Yet many correct completions lie so deep in the tail that they are effectively invisible under finite sampling. Training can increase their probability and make them accessible in practice while leaving nominal support unchanged. We refer to this movement from negligible toward observable probability as \emph{empirical support diffusion}. This motivates measuring accessibility at a positive probability resolution $\epsilon$. Our benchmarks record whether a question receives a verified solution, so we aggregate probability over its correct completions and apply the threshold at the question level under the stated decoding protocol.

\begin{definition}[Question-Level Solution Accessibility]
Let $\mathcal{D}_{\mathrm{eval}}$ be the evaluated set of questions. For a model distribution $p$, let $\widetilde p(y\mid x)$ denote the response distribution induced by its decoding protocol, including temperature and any truncation. Define the verified-success probability
\[
s_p(x) = \Pr_{y\sim\widetilde p(\cdot\mid x)}[R(x,y)=1].
\]
At resolution $\epsilon>0$, the accessible-question set is
\[
A_\epsilon(p)=\{x\in\mathcal{D}_{\mathrm{eval}}:s_p(x)>\epsilon\}.
\]

\end{definition}

\paragraph{Empirical Support under Finite Sampling.}
\label{app:soft_support}
Our benchmark tables compare observed success sets. For $n$ responses $y_1,\ldots,y_n$ sampled from $\widetilde p(\cdot\mid x)$, let
\[
\widehat A_n(p)=\left\{x\in\mathcal{D}_{\mathrm{eval}}:\sum_{i=1}^n R(x,y_i)>0\right\}.
\]
Thus, a question enters empirical support when at least one response passes verification. Under independent sampling, its probability of inclusion is $1-(1-s_p(x))^n$. The sampling budget therefore determines how reliably rare successes are detected: questions with $s_p(x)$ far below $1/n$ are rarely observed. We compare checkpoints at matched sampling budgets and decoding settings using the same benchmark acceptance criterion. Appx.~\ref{appendix:threshold} gives the success-probability bound after zero observed successes.

\subsection{Characterizing Solution Access Change}

With empirical support, we categorize changes in observed question-level success under RLVR:
\begin{definition}[Empirical Support Dynamics]
At a fixed evaluation budget and protocol, write $A=\widehat A_n(q)$ and $B=\widehat A_n(\pi_\theta)$. Questions in $A\cap B$ form \emph{preservation}: both models produce a verified success. Those in $B\setminus A$ form \emph{expansion}, where only the RLVR model succeeds, while those in $A\setminus B$ form \emph{shrinkage}, where only the reference model succeeds. Questions in $\mathcal{D}_{\mathrm{eval}}\setminus(A\cup B)$ are \emph{out of support}: neither model produces an observed verified success. Let $P$, $E$, $S$, and $O$ be the corresponding question counts.
\end{definition}

\paragraph{Support Dynamics Metrics.} To quantify RLVR's impact on solution accessibility, we use precision- and recall-inspired metrics for the four observed support categories.
\begin{definition}[Support Dynamics Metrics]
Let $P$, $E$, $S$, and $O$ denote the question counts in preservation, expansion, shrinkage, and out-of-support, respectively.

\textbf{Support Retention Rate (SRR)} measures retention of the reference model's observed successes:
\(\text{SRR}=\frac{P}{P+S}\in[0,1]\). SRR $=1$ means no observed shrinkage.

\textbf{New Discovery Rate (NDR)} measures the fraction of the RLVR model's observed successes that are newly observed relative to the reference:
\(\text{NDR}=\frac{E}{P+E}\in[0,1]\). NDR $=0$ means no observed expansion; NDR $=1$ means all observed RLVR successes are expansions.

\textbf{Support Dynamic Score (SDS)} is the harmonic mean of retention and discovery:
\[
\begin{aligned}
\text{SDS}&=\frac{2\cdot\text{SRR}\cdot\text{NDR}}{\text{SRR}+\text{NDR}}\\
&=\frac{2PE}{P^2+2PE+ES}\in[0,1].
\end{aligned}
\]
High SDS requires both retention and expansion.

\textbf{Net Support Change Rate (NSCR)} measures net observed support change:
\(\text{NSCR}=\frac{E-S}{P+E+S}\). Positive values indicate net expansion; negative values indicate net shrinkage.
Rates with zero denominators are undefined. When SRR and NDR are both defined, SDS is zero if either is zero.
\end{definition}
Read these summaries with the four counts. High retention with little discovery indicates preservation-dominated support. Large expansion and shrinkage counts reveal turnover even when NSCR is close to zero; net coverage grows when $E>S$. The opportunity for discovery also depends on the reference model: if it already solves nearly every evaluated question, few remain to be newly solved. With high retention, this limits both NDR and SDS. Appx.~\ref{sec:theory_rlvr} derives the effect of reward reweighting and the probability resolution of finite sampling.

\section{Empirical Support Dynamics across Models and Tasks}
\begin{table*}[!t]
\centering
\caption{Aggregate observed support dynamics. Each question is classified by whether at least one response passes verification for both checkpoints ($P$), only the trained checkpoint ($E$), only the reference checkpoint ($S$), or neither ($O$), at the benchmark-specific sampling budgets. SRR summarizes retention, NDR the fraction of trained-model successes newly observed relative to the reference, and SDS their harmonic mean. Positive NSCR denotes net expansion; negative NSCR denotes net shrinkage. Read the rates with the four counts. \texttt{Kangheng-OVR-7B} is a vision-language model (VLM). Per-benchmark counts and budgets appear in Appx.~\ref{app:support_across_models}.}
\label{tab:cross_model_comparison}
\begingroup
\small
\begin{tabular}{@{}c|c|cccc|cccc@{}}
\toprule
\rowcolor{blue!15}
\textbf{Model} & \textbf{Domain} & \textbf{SRR} & \textbf{NDR} & \textbf{SDS} & \textbf{NSCR} & 
{\textbf{P}} & \textcolor{MyGreen}{\textbf{E}} & \textcolor{red}{\textbf{S}} & \textcolor{gray}{\textbf{O}} \\
\midrule
\multirow{3}{*}{\textsc{ProRL-1.5B-v1}} &  Math     & 0.96 & 0.00 & 0.01 & -0.04 & 1355 & 5  & 56  & 131 \\
&  Non-Math & 0.91 & 0.03 & 0.06 & -0.06 & 1045 & 31 & 107 & 674 \\
&  Overall  & 0.94 & 0.01 & 0.03 & -0.05 & 2400 & 36 & 163 & 805 \\
\midrule 
\multirow{3}{*}{\textsc{ProRL-1.5B-v2}} &  Math     & 0.96 & 0.01 & 0.01 & -0.04 & 1349 & 9  & 62  & 127 \\
&  Non-Math & 0.90 & 0.04 & 0.07 & -0.06 & 1039 & 39 & 113 & 666 \\
&  Overall  & 0.93 & 0.02 & 0.04 & -0.05 & 2388 & 48 & 175 & 793 \\
\midrule
\multirow{3}{*}{\textsc{Nemotron-1-7B}} &  Math     & 0.99 & 0.00 & 0.01 & -0.00 & 1431 & 5  & 9  & 102 \\
&  Non-Math & 0.96 & 0.02 & 0.03 & -0.02 & 1284 & 23 & 47 & 503 \\
&  Overall  & 0.98 & 0.01 & 0.02 & -0.01 & 2715 & 28 & 56 & 605 \\
\midrule
\multirow{3}{*}{\textsc{Skywork-OR1-7B}} &  Math     & 0.98 & 0.00 & 0.00 & -0.02 & 1406 & 2  & 34 & 105 \\
&  Non-Math & 0.96 & 0.02 & 0.04 & -0.02 & 1279 & 24 & 52 & 502 \\
&  Overall  & 0.97 & 0.01 & 0.02 & -0.02 & 2685 & 26 & 86 & 607 \\
\midrule
\multirow{3}{*}{\textsc{Nemotron-1-14B}} &  Math     & 0.99 & 0.00 & 0.01 & -0.01 & 1425 & 5  & 15 & 102 \\
&  Non-Math & 0.99 & 0.00 & 0.01 & -0.00 & 993  & 3  & 8  & 399 \\
&  Overall  & 0.99 & 0.00 & 0.01 & -0.01 & 2418 & 8  & 23 & 501 \\
\midrule
\multirow{3}{*}{\textsc{Phi4-Reason-Plus-14B}} &  Math     & 0.99 & 0.01 & 0.01 & -0.00 & 1407 & 8  & 12 & 120 \\
&  Non-Math & 0.99 & 0.01 & 0.01 & -0.00 & 1067 & 8  & 11 & 317 \\
&  Overall  & 0.99 & 0.01 & 0.01 & -0.00 & 2474 & 16 & 23 & 437 \\
\midrule
{\textsc{OLMo-2-0425-1B}} & {Math}  & 0.88 & 0.10 & 0.18 &-0.02 & {761} & {83} & {104} & {599} \\
\midrule\textsc{Kangheng-OVR-7B} &  Math     & 0.99 & 0.00 & 0.01 & -0.00 & 781 & 3 & 4 & 516 \\ 
\bottomrule
\end{tabular}
\endgroup
\end{table*}

\subsection{Experimental Setup}
We adopt \textbf{ProRL-1.5B-v1}~\citep{liu2025prorl} as our main RLVR method due to its robust long-horizon training framework. Starting from \texttt{DeepSeek-R1-Distill-Qwen-1.5B} as the base model, ProRL’s \texttt{Nemotron-Research-Reasoning} series leverages GRPO enhanced with decoupled clipping, dynamic sampling, KL divergence regularization, and periodic reference resets to sustain exploration and mitigate entropy collapse during extended RL training. In addition, we evaluate other RLVR variants at multiple scales (7B–14B parameters), including \texttt{Skywork}~\citep{wei2023skywork}, \texttt{AceReason-Nemotron}~\citep{chen2025acereason}, and \texttt{Phi4-Reason}~\citep{abdin2025phi}, alongside a visual LLM (\texttt{Kangheng-OVR-7B}~\citep{wei2025open}). We additionally include ProRL-1.5B-v2 as a separate checkpoint for comparison (Tab.~\ref{tab:cross_model_comparison}).

Performance is measured across two categories. (1) \textbf{Math reasoning tasks}: MATH500~\citep{hendrycks2021measuring}, Minerva~\citep{lewkowycz2022solving}, OlympiadBench~\citep{he2024olympiadbench}, AIME 2024, AIME 2025, and AMC 2023. (2) \textbf{Non-math reasoning tasks}: SimpleQA~\citep{wei2024measuring} (factuality), LiveBench~\citep{white2025livebench} (logic, coding, and language comprehension), SciBench~\citep{wang2023scibench} (multi-domain scientific problem-solving), LiveCodeBench~\citep{jain2024livecodebench} (code generation), and Reasoning Gym~\citep{stojanovski2025reasoninggymreasoningenvironments} (cognition, geometry, graph theory, games). The VLM evaluation uses the MathVision~\citep{mathvision} and MathVista~\citep{mathvista} testmini sets. In Reasoning Gym, we especially focus on tasks that ProRL explicitly highlighted as challenging for the base model. For SimpleQA, we employ GPT-4.1 as the judge. Sampling budgets are $k\in\{4096,8192\}$ for the 1.5B--14B math comparisons, $k\in\{1024,2048,4096,8192,16384\}$ for Reasoning Gym, and $k\in\{1024,2048\}$ for other non-math datasets. The OLMo, SFT/DAPO, and training-checkpoint comparisons use $k=256$. Larger budgets make rare verified successes easier to detect. Appx.~\ref{appendix:threshold} gives a confidence bound on verified-success probability after zero observed successes. More implementation details appear in Appx.~\ref{app:exp}.

\begin{figure*}[!t]
    \centering
    \includegraphics[width=.85\textwidth]{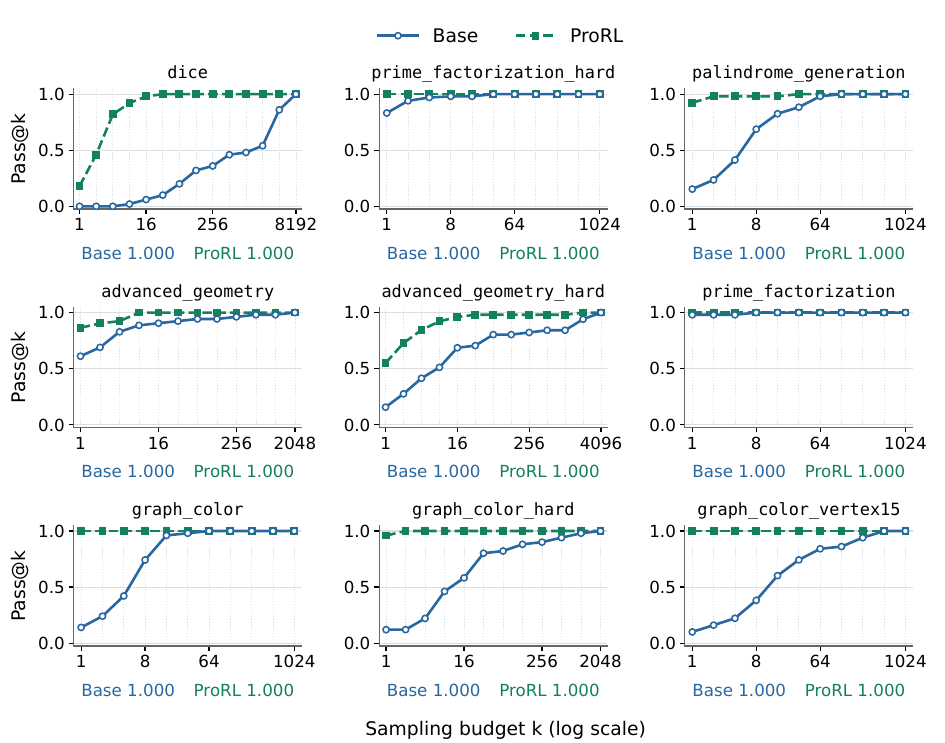}
    \vspace{-1em}
    \caption{{Empirical support preservation} across nine Reasoning Gym tasks. ProRL reaches high coverage with fewer samples, while the base model catches up as $k$ grows. Both models reach full coverage at the largest displayed budgets. Values below each panel report pass@$k$ at that budget; values for all budgets appear in Appx.~\ref{app:reasoning_gym_values}.}
    \label{fig:soft_support_in}
\end{figure*}
\begin{figure*}[!t]
    \centering
    \includegraphics[width=.85\textwidth]{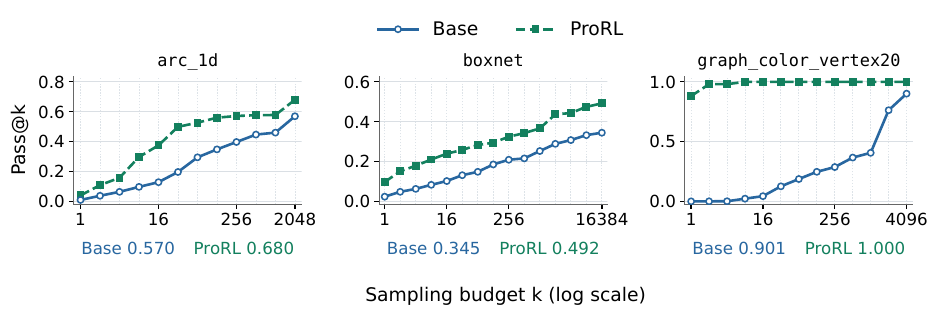}
    \vspace{-1em}
    \caption{Instances of \textcolor{MyGreen}{empirical support expansion} on \texttt{arc\_1d}, \texttt{boxnet}, and \texttt{graph\_color\_vertex20}. ProRL retains a coverage advantage through the largest sampling budget shown in each panel. Values below each panel report pass@$k$ at the largest displayed budget. Values for all budgets appear in Appx.~\ref{app:reasoning_gym_values}.}
    \label{fig:soft_support_out}
\end{figure*}
\begin{figure*}[!t]
    \centering
    \includegraphics[width=.85\textwidth]{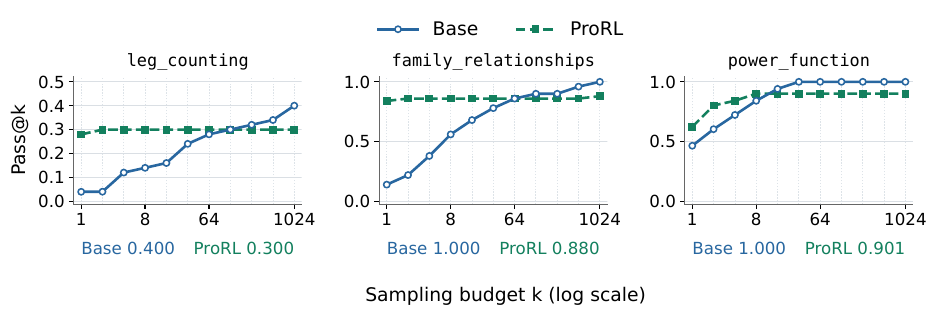}
    \vspace{-1em}
    \caption{\textcolor{red}{Empirical support shrinkage} on \texttt{leg\_counting}, \texttt{family\_relationships}, and \texttt{power\_function}. ProRL leads with one attempt, but the base model reaches greater coverage at larger sampling budgets. Values below each panel report pass@$k$ at the largest displayed budget; values for all budgets appear in Appx.~\ref{app:reasoning_gym_values}.}
    \label{fig:soft_support_shrinkage}
\end{figure*}
\subsection{Results: More Reliable Answers with Limited Expansion}

\paragraph{Large gains within preserved support.}
Across the six 1.5B--14B model comparisons, RLVR retains most questions the base model can solve (overall $\mathrm{SRR}\approx0.93$--$0.99$), while newly solved questions make up at most $2\%$ of RLVR successes ($\mathrm{NDR}\le0.02$; Tab.~\ref{tab:cross_model_comparison}). The Reasoning Gym curves show how preservation can accompany large accuracy gains: on hard graph coloring, ProRL answers $96\%$ of questions correctly in one attempt, compared with $12\%$ for the base model (Fig.~\ref{fig:soft_support_in}). With $2{,}048$ attempts per question, both reach $100\%$. RLVR makes correct answers much easier to obtain on questions already within the base model's reach.

\paragraph{Selective but limited empirical support expansion.}
RLVR also expands access on selected tasks. \texttt{ProRL-1.5B-v2} solves 48 questions on which the base model produces no verified success at the evaluation budget: 39 non-math and 9 math questions (Tab.~\ref{tab:cross_model_comparison}). These new questions account for $3.6\%$ of its non-math successes but only $0.7\%$ of its math successes ($\mathrm{NDR}=0.036$ versus $0.007$). ProRL-1.5B-v1 also retains a coverage advantage on \texttt{arc\_1d}, \texttt{boxnet}, and \texttt{graph\_color\_vertex20} through the largest sampling budget shown (Fig.~\ref{fig:soft_support_out}).

\paragraph{Empirical support shrinkage outweighs expansion.}
Losses exceed gains in every model-level aggregate in Tab.~\ref{tab:cross_model_comparison}, giving negative net support change ($\mathrm{NSCR}<0$). \texttt{ProRL-1.5B-v2} loses 175 questions while gaining 48, a ratio of approximately $3.6{:}1$. Even the 14B models show net shrinkage despite retaining nearly all base-model successes. This loss of coverage can remain hidden at small sampling budgets. On \texttt{leg\_counting}, ProRL succeeds on $28\%$ of questions in one attempt, compared with $4\%$ for the base model; at 1,024 attempts, the base model reaches $40\%$ while ProRL reaches only $30\%$. Related reversals occur on \texttt{family\_relationships} and \texttt{power\_function} (Fig.~\ref{fig:soft_support_shrinkage}). The high-budget deficit also appears on AIME2024, where base \texttt{pass@8192} reaches $93.3\%$, compared with \texttt{ProRL-1.5B-v1}'s $83.3\%$.

\noindent \textbf{Limited discovery keeps SDS low.}
Despite high retention, the six 1.5B--14B comparisons have SDS at most $0.07$ in each math and non-math aggregate: the harmonic mean remains low because newly solved questions make up only a small fraction of successes. \texttt{OLMo-2-0425-1B} has greater discovery ($\mathrm{NDR}=0.10$), raising SDS to $0.18$. Its 83 gains still fall short of 104 losses, however, so its higher SDS coexists with negative NSCR.

\noindent \textbf{The four support counts reveal turnover.}
Small net changes can also conceal changes in which questions are solved. In the AceReason-7B LiveBench-L comparison, both models solve 12 questions, yet only 10 are shared: two reference successes are lost and two new successes appear (Tab.~\ref{tab:support_nemotron}). Aggregate coverage stays at $24.0\%$ while the identities of successful questions change.

\noindent \textbf{Perplexity analysis of support changes.}
We examine how support changes relate to the likelihood of reasoning traces from DeepSeek-R1 and Sonnet 4 (Tab.~\ref{tab:combined_ppl}). In every reported shrinkage comparison, ProRL assigns lower likelihood to these traces; Sonnet 4 perplexity on AIME2024, for example, rises from 1.70 to 2.12. Perplexity also increases for questions missed by both models, rising from 8.76 to 14.91 on AIME2024 Sonnet 4 traces. Expansion reverses this pattern: Olympiad perplexity falls from 1.41 to 1.28 for DeepSeek-R1 traces and from 1.65 to 1.38 for Sonnet 4 traces, linking newly observed successes to greater likelihood of the supplied reasoning traces.

\textbf{Comparison between SFT and RLVR.} We fix the base model (Qwen2.5-Math-7B) and dataset (DeepMath-103K) to compare SFT with DAPO (Tab.~\ref{tab:combined_sft_dapo}). SFT maintains or increases total coverage on all six benchmarks (nonnegative NSCR), whereas DAPO produces net shrinkage on four. DAPO retains most base-model successes, but newly solved questions make up a smaller share of its successes than under SFT on five benchmarks (lower NDR). At small sampling budgets, DAPO nevertheless outperforms SFT on several tasks; with more samples, SFT's coverage advantage becomes visible. On Minerva, DAPO leads SFT with one attempt ($40.81\%$ versus $36.03\%$), but SFT leads with 256 attempts ($66.91\%$ versus $55.15\%$; Appx.~\ref{app:sft_dapo_passk}). The SFT comparison demonstrates that broader coverage is attainable from the same starting model and dataset.
\begin{table}[t]
\centering
% \vspace{-1em}
\caption{Reasoning-token perplexity of traces from DeepSeek-R1 and Sonnet 4, evaluated under the base model and ProRL. Questions are grouped by whether each model produces a verified success: only the base model (\textcolor{red}{shrinkage}), only ProRL (\textcolor{MyGreen}{expansion}), or neither. Higher perplexity indicates lower likelihood of the supplied reasoning traces.}
\label{tab:combined_ppl}
\resizebox{\columnwidth}{!}{%
\begin{tabular}{@{}ccc|c|ccc@{}} 
\toprule
\textbf{Category} & \textbf{Correctness} & \textbf{Reference} & \textbf{Target} & \textbf{AIME24} & \textbf{AIME25} & \textbf{Olympiad} \\ \midrule
 \multirow{4}{*}{\textcolor{red}{Shrinkage}} & \multirow{4}{*}{\ding{51} Base, \ding{55} ProRL} & \multirow{2}{*}{DeepSeek-R1} & Base  & 1.24 & 1.39 & 1.17  \\ 
 & &  & ProRL & 1.39 & 1.70 & 1.25 \\    \cmidrule{3-7}
 & & \multirow{2}{*}{Claude Sonnet 4} & Base  & 1.70 & 1.54 & 1.51  \\
 && & ProRL & 2.12 & 1.98 & 1.83 \\   \midrule 
\multirow{4}{*}{\textcolor{MyGreen}{Expansion}} & \multirow{4}{*}{\ding{55} Base, \ding{51} ProRL} & \multirow{2}{*}{DeepSeek-R1} & Base  & - & - & 1.41  \\
 & & & ProRL &  - & - & 1.28  \\   \cmidrule{3-7}
  & & \multirow{2}{*}{Claude Sonnet 4} & Base  &  - & - & 1.65\\
 && & ProRL & - & - & 1.38  \\  \midrule 
 \multirow{4}{*}{\textcolor{gray}{-- --}}& \multirow{4}{*}{\ding{55} Base, \ding{55} ProRL} & \multirow{2}{*}{DeepSeek-R1} & Base  & 1.82 & 1.75 & 1.62 \\
 & & & ProRL & 2.20 & 2.15 & 1.94 \\  \cmidrule{3-7}
 & & \multirow{2}{*}{Claude Sonnet 4} & Base  & 8.76 & 6.05 & 5.98 \\
 && & ProRL & 14.91 & 9.76 & 9.55 \\   \bottomrule
\end{tabular}}
% \vspace{-2em}
\end{table}

\begin{table*}[!t]
\centering
\caption{Observed support dynamics for SFT and DAPO training on {Qwen2.5-Math-7B}, evaluated at 256 samples per question. SFT has nonnegative net support change on all six benchmarks. DAPO has net expansion on AIME24 and AIME25 and net shrinkage on the other four benchmarks. Counts record question-level verified successes relative to the base model.}
\label{tab:combined_sft_dapo}
\begin{supporttable}
\fontsize{8}{9}\selectfont
\setlength{\tabcolsep}{3pt}
\renewcommand{\arraystretch}{0.90}
\begin{tabular}{@{}c|c|cc|cccc|cccc@{}} \toprule
\rowcolor{blue!15}
\multirow{2}{*}{\textbf{Dataset}} 
& \multirow{2}{*}{\textbf{Training}} 
& \multicolumn{2}{c|}{\textbf{pass@k}} 
& \multicolumn{4}{c|}{\textbf{Support Dynamics Metrics}} 
& \multicolumn{4}{c}{\textbf{Support Counts}} \\
\cmidrule(lr){3-4} \cmidrule(lr){5-8} \cmidrule(lr){9-12}
\rowcolor{blue!15}
& & \textbf{Base} & \textbf{Target} 
& \textbf{SRR} & \textbf{NDR} & \textbf{SDS} & \textbf{NSCR}
& {\textbf{P}} & \textcolor{MyGreen}{\textbf{E}}
& \textcolor{red}{\textbf{S}} & \textcolor{gray}{\textbf{O}} \\
\midrule
\rowcolor{purple!10}
\multicolumn{12}{c}{\textbf{\textit{Math Reasoning Benchmarks (pass@256)}}} \\
\midrule
\multirow{2}{*}{AIME24}
& SFT  & 63.33\% & 63.33\% & 0.789 & 0.211 & 0.332 & 0.000 & 15 & 4 & 4 & 7 \\
& DAPO & 63.33\% & 66.67\% & 0.895 & 0.150 & 0.257 & 0.045 & 17 & 3 & 2 & 8 \\
\midrule
\multirow{2}{*}{AIME25}
& SFT  & 50.00\% & 63.33\% & 0.933 & 0.263 & 0.411 & 0.200 & 14 & 5 & 1 & 10 \\
& DAPO & 50.00\% & 56.67\% & 1.000 & 0.118 & 0.211 & 0.118 & 15 & 2 & 0 & 13 \\
\midrule
\multirow{2}{*}{AMC23} & SFT  & 100.00\% & 100.00\% & 1.000 & 0.000 & 0.000 & 0.000 & 40 & 0 & 0 & 0 \\
& DAPO & 100.00\% & 92.50\%  & 0.925 & 0.000 & 0.000 & -0.075 & 37 & 0 & 3 & 0 \\
\midrule
\multirow{2}{*}{MATH500}
& SFT  & 96.40\% & 99.00\% & 0.996 & 0.030 & 0.059 & 0.026 & 480 & 15 & 2 & 3 \\
& DAPO & 96.40\% & 95.20\% & 0.975 & 0.013 & 0.025 & -0.012 & 470 & 6 & 12 & 12 \\
\midrule
\multirow{2}{*}{Minerva}
& SFT  & 63.24\% & 66.91\% & 0.895 & 0.154 & 0.263 & 0.050 & 154 & 28 & 18 & 72 \\
& DAPO & 63.24\% & 55.15\% & 0.802 & 0.080 & 0.145 & -0.120 & 138 & 12 & 34 & 88 \\
\midrule
\multirow{2}{*}{Olympiad}
& SFT  & 78.22\% & 81.78\% & 0.953 & 0.089 & 0.162 & 0.042 & 503 & 49 & 25 & 98 \\
& DAPO & 78.22\% & 72.89\% & 0.884 & 0.051 & 0.096 & -0.065 & 467 & 25 & 61 & 122 \\ \bottomrule
\end{tabular}
\end{supporttable}
% \vspace{-1em}
\end{table*}

\begin{table*}[!t]
\centering
\caption{{Effect of RLVR (DAPO) training on DeepSeek 1.5B across math benchmarks.} \texttt{Pass@256} values are percentages. Step = 0 corresponds to the non-RLVR baseline.}
\label{tab:support_training_process}
\begingroup
\small
\setlength{\tabcolsep}{3pt}
\renewcommand{\arraystretch}{0.80}
\resizebox{.86\textwidth}{!}{%
\begin{tabular}{l|c|c|cccc|cccc||l|c|c|cccc|cccc}
\toprule
\rowcolor{blue!15}
\multirow{2}{*}{\textbf{Dataset}} &
\multirow{2}{*}{\textbf{Step}} &
\textbf{Pass@256} &
\multicolumn{4}{c|}{\textbf{Support Dynamics Metrics}} &
\multicolumn{4}{c||}{\textbf{Support Counts}} &
\multirow{2}{*}{\textbf{Dataset}} &
\multirow{2}{*}{\textbf{Step}} &
\textbf{Pass@256} &
\multicolumn{4}{c|}{\textbf{Support Dynamics Metrics}} &
\multicolumn{4}{c}{\textbf{Support Counts}} \\
\cmidrule(lr){4-7} \cmidrule(lr){8-11} \cmidrule(lr){15-18} \cmidrule(lr){19-22}
\rowcolor{blue!15}
 & & \textbf{RLVR} & \textbf{SRR} & \textbf{NDR} & \textbf{SDS} & \textbf{NSCR} &
 {\textbf{P}} & \textcolor{MyGreen}{\textbf{E}} & \textcolor{red}{\textbf{S}} & \textcolor{gray}{\textbf{O}} &
 & & \textbf{RLVR} & \textbf{SRR} & \textbf{NDR} & \textbf{SDS} & \textbf{NSCR} &
 {\textbf{P}} & \textcolor{MyGreen}{\textbf{E}} & \textcolor{red}{\textbf{S}} & \textcolor{gray}{\textbf{O}} \\
\midrule
\rowcolors{3}{blue!4}{white}

% ================== AIME24  ||  AIME25 ==================
\multirow{11}{*}{AIME24}
  & 0   & 83.33 & -     & -     & -     & -      & -  & - & - & - &
\multirow{11}{*}{AIME25}
  & 0   & 70.00 & -     & -     & -     & -      & -  & - & - & - \\
  & 30  & 80.00 & 0.960 & 0.000 & 0.000 & -0.040 & 24 & 0 & 1 & 5 &
  & 30  & 70.00 & 0.905 & 0.095 & 0.172 &  0.000 & 19 & 2 & 2 & 7 \\
  & 60  & 80.00 & 0.960 & 0.000 & 0.000 & -0.040 & 24 & 0 & 1 & 5 &
  & 60  & 73.33 & 0.905 & 0.136 & 0.237 &  0.042 & 19 & 3 & 2 & 6 \\
  & 90  & 83.33 & 1.000 & 0.000 & 0.000 &  0.000 & 25 & 0 & 0 & 5 &
  & 90  & 70.00 & 0.905 & 0.095 & 0.172 &  0.000 & 19 & 2 & 2 & 7 \\
  & 120 & 80.00 & 0.960 & 0.000 & 0.000 & -0.040 & 24 & 0 & 1 & 5 &
  & 120 & 73.33 & 0.905 & 0.136 & 0.237 &  0.042 & 19 & 3 & 2 & 6 \\
  & 150 & 80.00 & 0.960 & 0.000 & 0.000 & -0.040 & 24 & 0 & 1 & 5 &
  & 150 & 73.33 & 0.905 & 0.136 & 0.237 &  0.042 & 19 & 3 & 2 & 6 \\
  & 180 & 83.33 & 1.000 & 0.000 & 0.000 &  0.000 & 25 & 0 & 0 & 5 &
  & 180 & 63.33 & 0.905 & 0.000 & 0.000 & -0.095 & 19 & 0 & 2 & 9 \\
  & 210 & 76.67 & 0.920 & 0.000 & 0.000 & -0.080 & 23 & 0 & 2 & 5 &
  & 210 & 66.67 & 0.905 & 0.050 & 0.095 & -0.045 & 19 & 1 & 2 & 8 \\
  & 240 & 73.33 & 0.880 & 0.000 & 0.000 & -0.120 & 22 & 0 & 3 & 5 &
  & 240 & 66.67 & 0.905 & 0.050 & 0.095 & -0.045 & 19 & 1 & 2 & 8 \\
  & 270 & 73.33 & 0.880 & 0.000 & 0.000 & -0.120 & 22 & 0 & 3 & 5 &
  & 270 & 63.33 & 0.905 & 0.000 & 0.000 & -0.095 & 19 & 0 & 2 & 9 \\
  & 300 & 73.33 & 0.880 & 0.000 & 0.000 & -0.120 & 22 & 0 & 3 & 5 &
  & 300 & 66.67 & 0.857 & 0.100 & 0.179 & -0.043 & 18 & 2 & 3 & 7 \\
\midrule

% ================== AMC23  ||  MATH500 ==================
\multirow{11}{*}{AMC23}
  & 0   & 97.50 & -     & -     & -     & -     & -  & - & - & - &
\multirow{11}{*}{MATH500}
  & 0   & 99.20 & -     & -     & -     & -      & -   & - & - & - \\
  & 30  & 97.50 & 1.000 & 0.000 & 0.000 & 0.000 & 39 & 0 & 0 & 1 &
  & 30  & 98.80 & 0.996 & 0.000 & 0.000 & -0.004 & 494 & 0 & 2 & 4 \\
  & 60  & 97.50 & 1.000 & 0.000 & 0.000 & 0.000 & 39 & 0 & 0 & 1 &
  & 60  & 98.80 & 0.996 & 0.000 & 0.000 & -0.004 & 494 & 0 & 2 & 4 \\
  & 90  & 97.50 & 1.000 & 0.000 & 0.000 & 0.000 & 39 & 0 & 0 & 1 &
  & 90  & 99.20 & 0.998 & 0.002 & 0.004 &  0.000 & 495 & 1 & 1 & 3 \\
  & 120 & 97.50 & 1.000 & 0.000 & 0.000 & 0.000 & 39 & 0 & 0 & 1 &
  & 120 & 99.00 & 0.996 & 0.002 & 0.004 & -0.002 & 494 & 1 & 2 & 3 \\
  & 150 & 97.50 & 1.000 & 0.000 & 0.000 & 0.000 & 39 & 0 & 0 & 1 &
  & 150 & 98.80 & 0.996 & 0.000 & 0.000 & -0.004 & 494 & 0 & 2 & 4 \\
  & 180 & 97.50 & 1.000 & 0.000 & 0.000 & 0.000 & 39 & 0 & 0 & 1 &
  & 180 & 99.00 & 0.998 & 0.000 & 0.000 & -0.002 & 495 & 0 & 1 & 4 \\
  & 210 & 97.50 & 1.000 & 0.000 & 0.000 & 0.000 & 39 & 0 & 0 & 1 &
  & 210 & 98.80 & 0.994 & 0.002 & 0.004 & -0.004 & 493 & 1 & 3 & 3 \\
  & 240 & 97.50 & 1.000 & 0.000 & 0.000 & 0.000 & 39 & 0 & 0 & 1 &
  & 240 & 98.60 & 0.992 & 0.002 & 0.004 & -0.006 & 492 & 1 & 4 & 3 \\
  & 270 & 97.50 & 1.000 & 0.000 & 0.000 & 0.000 & 39 & 0 & 0 & 1 &
  & 270 & 98.40 & 0.992 & 0.000 & 0.000 & -0.008 & 492 & 0 & 4 & 4 \\
  & 300 & 97.50 & 1.000 & 0.000 & 0.000 & 0.000 & 39 & 0 & 0 & 1 &
  & 300 & 97.80 & 0.984 & 0.002 & 0.004 & -0.014 & 488 & 1 & 8 & 3 \\
\midrule

% ================== Minerva  ||  Olympiad ==================
\multirow{11}{*}{Minerva}
  & 0   & 62.50 & -     & -     & -     & -      & -   & -  & -  & -  &
\multirow{11}{*}{Olympiad}
  & 0   & 88.00 & -     & -     & -     & -      & -   & -  & -  & -  \\
  & 30  & 62.13 & 0.935 & 0.059 & 0.111 & -0.006 & 159 & 10 & 11 & 92 &
  & 30  & 86.67 & 0.963 & 0.022 & 0.043 & -0.015 & 572 & 13 & 22 & 68 \\
  & 60  & 59.93 & 0.929 & 0.031 & 0.059 & -0.040 & 158 & 5  & 12 & 97 &
  & 60  & 86.81 & 0.960 & 0.027 & 0.053 & -0.013 & 570 & 16 & 24 & 65 \\
  & 90  & 61.76 & 0.924 & 0.065 & 0.122 & -0.011 & 157 & 11 & 13 & 91 &
  & 90  & 86.07 & 0.961 & 0.017 & 0.034 & -0.022 & 571 & 10 & 23 & 71 \\
  & 120 & 62.87 & 0.953 & 0.053 & 0.100 &  0.006 & 162 & 9  & 8  & 93 &
  & 120 & 86.52 & 0.961 & 0.022 & 0.044 & -0.016 & 571 & 13 & 23 & 68 \\
  & 150 & 62.50 & 0.947 & 0.053 & 0.100 &  0.000 & 161 & 9  & 9  & 93 &
  & 150 & 85.48 & 0.955 & 0.017 & 0.034 & -0.028 & 567 & 10 & 27 & 71 \\
  & 180 & 59.56 & 0.906 & 0.049 & 0.094 & -0.045 & 154 & 8  & 16 & 94 &
  & 180 & 84.00 & 0.944 & 0.011 & 0.021 & -0.045 & 561 & 6  & 33 & 75 \\
  & 210 & 60.29 & 0.924 & 0.043 & 0.082 & -0.034 & 157 & 7  & 13 & 95 &
  & 210 & 82.81 & 0.926 & 0.016 & 0.032 & -0.058 & 550 & 9  & 44 & 72 \\
  & 240 & 61.40 & 0.941 & 0.042 & 0.080 & -0.017 & 160 & 7  & 10 & 95 &
  & 240 & 82.81 & 0.931 & 0.011 & 0.021 & -0.058 & 553 & 6  & 41 & 75 \\
  & 270 & 59.19 & 0.906 & 0.043 & 0.083 & -0.051 & 154 & 7  & 16 & 95 &
  & 270 & 81.33 & 0.912 & 0.013 & 0.025 & -0.075 & 542 & 7  & 52 & 74 \\
  & 300 & 58.46 & 0.912 & 0.025 & 0.049 & -0.063 & 155 & 4  & 15 & 98 &
  & 300 & 79.70 & 0.897 & 0.009 & 0.018 & -0.093 & 533 & 5  & 61 & 76 \\
\bottomrule
\end{tabular}
}
\endgroup
% \vspace{-1em}
\end{table*}
\noindent \textbf{Support dynamics during RLVR training.} We track DAPO training on DeepSeek-1.5B to examine how empirical support evolves across checkpoints (Tab.~\ref{tab:support_training_process}). Most base-model successes remain accessible throughout training (high SRR), yet newly solved questions fail to offset the losses on five of the six benchmarks by step 300 (negative NSCR). On Olympiad, gains decrease from 13 at step 30 to 5 at step 300, while losses increase from 22 to 61. This widening imbalance leaves final \texttt{pass@256} at $79.70\%$, below the initial $88.00\%$. On the harder AIME25 and Minerva, intermediate checkpoints achieve net expansion, but both finish with net shrinkage at step 300. Such losses of earlier successes are also reported for both RL and SFT in the temporal-forgetting study of \citet{li2025temporal}.

\noindent \textbf{Takeaway: reliability gains with net support loss.}
RLVR makes existing successes easier to sample in the preservation examples and adds successes on selected tasks. Yet losses exceed gains in every reported model-level aggregate across the evaluated 1B--14B models, math and non-math tasks, and the vision-language comparison. Intermediate expansion can also disappear with continued training. These motivate exploration strategies that make successful responses in low-probability regions easier to reach while retaining existing successes.

\section{Token Uncertainty and Answer Concentration}

\subsection{Experimental Setup}
Prior work links high-entropy tokens to branching decisions in reasoning~\citep{wang2025beyond} and reports sharp declines in policy entropy early in RLVR training~\citep{cui2025entropy}. We examine whether greater uncertainty at individual generation steps corresponds to more varied final answers. For each reference and RLVR model, we sample $N=32$ responses per question on the math benchmarks and report \texttt{avg@32} accuracy, the proportion of correct responses averaged over questions, together with two entropy measures:
\begin{itemize}[left=0pt]
    \item \textbf{Token-Level Entropy:}  
    Let $\mathcal{V}$ denote the vocabulary and $y^{(i)} = (y^{(i)}_1, y^{(i)}_2, \ldots, y^{(i)}_{T^{(i)}})$ denote the $i$-th generated sequence of length $T^{(i)}$ for $1\leq i \leq N$. At each timestep $t$, the model outputs a probability distribution $p_t^{(i)}(v)$ over vocabulary tokens $v \in \mathcal{V}$. The entropy of this distribution is given by:
    $H_t^{(i)} = -\sum_{v \in \mathcal{V}} p_t^{(i)}(v) \log p_t^{(i)}(v)$. The average token-level entropy over all $N$ sequences and their timesteps is computed as:
    \(\text{TokenEntropy} = \frac{1}{N} \sum_{i=1}^{N} \left( \frac{1}{T^{(i)}} \sum_{t=1}^{T^{(i)}} H_t^{(i)} \right)\), capturing the local uncertainty at each generation step.

    \item \textbf{Answer-level Entropy:}  
    Let $\{o^{(1)}, \ldots, o^{(N)}\}$ denote the answers extracted from each generated sequence $y^{(i)}$ (using \textsc{NA} for incomplete outputs), and let $\{o_1^\ast, \ldots, o_M^\ast\}$ be the $M$ unique answers. Let $f_j$ be the frequency of answer $o_j^\ast$, with empirical probability $p_j = \frac{f_j}{N}$. Then:
    \(\text{AnswerEntropy} = -\sum_{j=1}^{M} p_j \log p_j\). 
    This finite-sample estimate measures how evenly the 32 responses are distributed across extracted final answers, including incorrect answers and \textsc{NA}. Lower values indicate greater concentration on particular answers.
\end{itemize}

\subsection{Accuracy Gains and Entropy Dynamics} 

\paragraph{Accuracy gains accompany greater answer concentration.}
RLVR improves \texttt{avg@32} accuracy and reduces answer-level entropy in every model--benchmark comparison in Tab.~\ref{tab:entropy}. For ProRL, average accuracy rises from $54.5\%$ to $65.4\%$ as answer entropy falls from $1.30$ to $0.66$, indicating that more accurate responses accompany a sharper final-answer distribution. DAPO-32B shows the same pattern: accuracy rises from $43.0\%$ to $61.3\%$ while answer entropy falls from $1.61$ to $0.61$. Answer entropy also declines on the harder AIME24 and Minerva benchmarks, although both reference and RLVR models remain more uncertain there than on AMC23 and MATH500.

% \vspace{-0.5em}

\paragraph{Decoupled local uncertainty and answer concentration.}
While answer-level entropy consistently declines, token-level entropy changes in different directions across models. ProRL's average token entropy rises from $0.44$ to $0.52$, and DAPO-32B's rises from $0.18$ to $0.29$. Both models have greater uncertainty at individual generation steps even as their final answers become more concentrated. By contrast, both AceReason models and Skywork reduce entropy at both levels relative to their reference checkpoints.

The same distinction appears between models sharing a reference checkpoint. Relative to AceReason-7B, Skywork-OR1-7B has lower token entropy ($0.16$ versus $0.23$) but higher answer entropy ($0.54$ versus $0.50$), with this ordering holding on all five benchmarks. Here, the model with less local uncertainty produces the less concentrated answer distribution.

\paragraph{Implications.} For tasks with a unique correct answer, concentrating probability on that answer is useful. The ProRL and DAPO-32B results also show that higher token entropy can accompany more concentrated final answers, so local uncertainty alone gives an incomplete account of exploration. The support analysis identifies a complementary issue: newly solved questions can be outweighed by lost successes. An effective exploration objective should broaden verified solution access while preserving prior successes.

% \vspace{-0.3em}
\section{Conclusion}
We introduce empirical support dynamics to quantify what RLVR preserves, gains, and loses in access to verified solutions. Across the model-level benchmark aggregates, most base-model successes remain available, yet losses exceed gains even as single-sample accuracy improves. ProRL and DAPO-32B also exhibit higher token-level entropy alongside more concentrated final answers, so increased local uncertainty alone does not establish broader answer diversity. Future RLVR should be judged by whether they bring additional questions into reach, retain prior successes, and sustain newly gained access through continued training. % To extend reasoning beyond the base model's scope, RLVR must be paired with explicit exploration strategies or off-policy mechanisms that allocate probability mass to underrepresented regions of the solution space.

% \section*{Impact Statements}
% This work examines the strengths and limitations of reinforcement learning with verifiable rewards (RLVR) in large language models, focusing on how such training reshapes the accessibility of correct solutions rather than introducing new capabilities. By providing empirical evidence that RLVR primarily improves precision while potentially narrowing solution diversity, our findings aim to help researchers and practitioners better understand the trade-offs involved in deploying RLVR-based systems. We do not anticipate direct negative societal impacts from this analysis; instead, we hope it encourages more transparent evaluation practices and motivates the development of training methods that balance reliability with broader reasoning coverage.
 
% \section*{Acknowledgment}
% This work was supported in part by RS-2024-00457882, National AI Research Lab Project. This work was supported by IITP funded by the Korean Government (MSIT) (No. RS-2024-00457882, National AI Research Lab Project).  This work was supported by the Institute of Information & Communications Technology Planning & Evaluation (IITP) grant funded by the Korean Government (MSIT) (No. RS-2024-00457882, National AI Research Lab Project).

% We thank Xu Huang (Georgia Institute of Technology) for insightful suggestions that strengthened the presentation of this work.

\section*{Limitations}
We evaluate the listed checkpoints and training recipes under the stated decoding settings and sampling budgets. Finite sampling makes membership in the observed support set variable across repeated draws, especially for questions with rare successful responses. Verification follows each benchmark's acceptance criterion, which may evaluate final answers without checking the intervening reasoning~\citep{wen2025reinforcement}. At large sampling budgets, even infrequent false acceptances can affect observed support. Our comparisons measure the outcomes of complete training recipes; identifying which components cause expansion or shrinkage requires matched interventions.

\bibliography{cite}

\clearpage
\appendix

\section{Related Works}

RLVR has emerged as a scalable alternative to RLHF by replacing human preference labels with automated, verifiable signals. Unlike RLHF, which relies on learned reward models for subjective alignment, RLVR optimizes objective, algorithmically computable rewards that directly reflect ground-truth success criteria in domains with clear verifiability. It has achieved strong gains on real-world applications, including math~\citep{guo2025deepseek} and medical multiple-choice QA~\citep{zhang2025med,lai2025med,feng2025doctoragent}, often outperforming SFT with limited training data.

Recent analyses have examined both the benefits and limitations of RLVR. Studies~\citep{wen2025reinforcement} argue that standard evaluation metrics, such as Pass@K, can overstate RLVR gains by ignoring the correctness of intermediate reasoning, motivating alternatives (e.g., CoT-Pass@K) that jointly assess reasoning chains and final answers. Under such metrics, RLVR exhibits stronger incentives toward correct reasoning. Complementarily, a recent work by~\citet{AmitGPR25} presents an interesting theoretical perspective in which verifier-based training can provide per-instance correctness guarantees by conditioning acceptance on externally verifiable proofs, thereby shifting evaluation from average-case accuracy to worst-case soundness. A concurrent position highlights hidden costs in measurement protocols, data contamination, and evaluation bias, advocating tax-aware evaluation that jointly optimizes accuracy, grounding, and calibrated abstention~\citep{tu2025position}.

At the algorithmic level, recent RLVR variants address challenges in reward sparsity, verifier reliability, and exploration. \citet{cai2025reinforcement} explicitly models asymmetric noise in automated verifiers to improve robustness under imperfect rewards. Self-verification approaches~\citep{liu2025trust} integrate generation and critique within a single RL loop to produce denser and more informative feedback. Other work explores efficiency-oriented objectives and supervised surrogates to stabilize optimization under sparse verifiable signals~\citep{sun2025efficient}. Finally, emerging theoretical perspectives investigate semantic hidden-state dynamics beyond token-level entropy, proposing structural incentives such as VERL to better decouple exploration and exploitation in reasoning models~\citep{huang2025beyond}.

\section{Detailed SFT and DAPO Sampling Profiles}
\label{app:sft_dapo_passk}

Table~\ref{tab:sft_rlvr_passk} provides the detailed \texttt{pass@k} profiles for five math benchmarks. The paired question-level support comparison at 256 samples per question is reported in the main paper (Tab.~\ref{tab:combined_sft_dapo}).

\begin{table*}[t]
\centering
{\caption{Detailed \texttt{pass@k} comparison of Qwen2.5-Math-7B trained with SFT or DAPO across five math benchmarks. Values are percentages; column headings give the number of responses $k$. Higher values indicate greater observed verified coverage.}
\label{tab:sft_rlvr_passk}
\resizebox{0.68\textwidth}{!}{%
\begin{tabular}{lccccccccc} \toprule
\rowcolor{blue!15}\textbf{AIME2024} & @1 & @2 & @4 & @8 & @16 & @32 & @64 & @128 & @256 \\ \midrule
SFT  & 20.00 & 30.00 & 30.00 & 33.33 & 43.33 & 50.00 & 63.33 & 63.33 & 63.33 \\
DAPO & 20.00 & 26.67 & 26.67 & 40.00 & 46.67 & 56.67 & 60.00 & 63.33 & 66.67 \\ \midrule
\rowcolor{blue!15}\textbf{AIME2025} & @1 & @2 & @4 & @8 & @16 & @32 & @64 & @128 & @256 \\ \midrule
SFT  & 16.67 & 16.67 & 26.67 & 30.00 & 33.33 & 40.00 & 43.33 & 50.00 & 63.33 \\
DAPO & 16.67 & 20.00 & 30.00 & 43.33 & 46.67 & 50.00 & 50.00 & 50.00 & 56.67 \\ \midrule
\rowcolor{blue!15}\textbf{AMC23} & @1 & @2 & @4 & @8 & @16 & @32 & @64 & @128 & @256 \\ \midrule
SFT  & 60.00 & 72.50 & 77.50 & 85.00 & 90.00 & 95.00 & 97.50 & 97.50 & 100.00 \\
DAPO & 62.50 & 72.50 & 85.00 & 87.50 & 87.50 & 92.50 & 92.50 & 92.50 & 92.50 \\ \midrule
\rowcolor{blue!15}\textbf{MATH500} & @1 & @2 & @4 & @8 & @16 & @32 & @64 & @128 & @256 \\ \midrule
SFT  & 78.20 & 85.20 & 90.00 & 92.80 & 95.00 & 96.20 & 98.00 & 98.80 & 99.00 \\
DAPO & 85.00 & 89.80 & 91.40 & 93.00 & 93.20 & 93.80 & 94.40 & 94.80 & 95.20 \\ \midrule
\rowcolor{blue!15}\textbf{Minerva-Math} & @1 & @2 & @4 & @8 & @16 & @32 & @64 & @128 & @256 \\ \midrule
SFT  & 36.03 & 39.71 & 43.75 & 47.79 & 51.47 & 54.04 & 63.60 & 65.44 & 66.91 \\
DAPO & 40.81 & 45.96 & 48.16 & 51.10 & 51.10 & 52.57 & 53.68 & 54.41 & 55.15 \\ \bottomrule
\end{tabular}}}
\end{table*}

\section{Detailed Statistics for Support Dynamics}
\label{app:support_across_models}

We provide per-model statistics for the detailed comparisons reported below. These counts underlie the corresponding aggregate values in Tab.~\ref{tab:cross_model_comparison}. For each model and domain (Math, Non-Math, and Overall), we report question counts across the four observed support categories: \emph{Preservation (P)}, \emph{Expansion (E)}, \emph{Shrinkage (S)}, and \emph{Out-of-Support (O)}. A question enters a model's observed success set when at least one response passes verification at the stated budget. From these counts, we compute Support Retention Rate (SRR), New Discovery Rate (NDR), Support Dynamic Score (SDS), and Net Support Change Rate (NSCR).

We include detailed results for \texttt{ProRL-1.5B-v1}, \texttt{Nemotron-1-7B}, \texttt{Skywork-OR1-7B}, \texttt{Nemotron-1-14B}, \texttt{Phi4-Reason-Plus-14B}, \texttt{OLMo-2-0425-1B}, and the visual reasoning model \texttt{Kangheng-OVR-7B (VLM)}.

\subsection{Possible Sources of Empirical Support Expansion}

On \texttt{arc\_1d}, \texttt{boxnet}, and \texttt{graph\_color\_vertex20}, the base model rarely succeeds in a single attempt, yet its coverage rises as more responses are sampled (Fig.~\ref{fig:soft_support_out}). RLVR's additional gains on these tasks motivate two hypotheses about the reliability of producing complete, verifiable responses.

\paragraph{Composing partially reliable subskills.}
Several expanding tasks require coordinated operations: local assignments in graph coloring, element-wise transformations in \texttt{arc\_1d}, or JSON-style key--value fragments in \texttt{boxnet}. A model may produce useful individual operations yet rarely assemble a complete correct solution. Rewarding successful combinations could increase the reliability of their joint execution, making complete solutions easier to sample. Under this hypothesis, expansion arises from improved coordination of partially reliable subskills.

\paragraph{Aligning responses with the required format.}
A successful response must also express its solution in a form accepted by the verifier. Our Reasoning Gym evaluation encountered base-model answers requiring fraction normalization in \texttt{dice}, digit extraction and spacing in \texttt{arc\_1d}, and cleaning of JSON structures in \texttt{boxnet} (Appx.~\ref{app:reasonining_gym_evaluation}). These observations motivate a second hypothesis: RLVR can improve verified success by making otherwise useful responses conform more reliably to the required format. We measure expansion after applying these answer-processing rules.

\begin{table*}[t]
\centering
\caption{Support dynamics metrics and \texttt{pass@k} performance of ProRL-1.5B-v1, compared with its base model, DeepSeek-R1-Distill-Qwen-1.5B.}
\label{tab:support_prorl}
\begin{supporttable}
\begin{tabular}{@{}c|cc|cccc|cccc@{}} \toprule
\rowcolor{blue!15}
\multirow{2}{*}{\textbf{Dataset}} & \multicolumn{2}{c|}{\textbf{pass@k Performance}} & \multicolumn{4}{c|}{\textbf{Support Dynamics Metrics}} & \multicolumn{4}{c}{\textbf{Support Counts}} \\
\cmidrule(lr){2-3} \cmidrule(lr){4-7} \cmidrule(lr){8-11}
\rowcolor{blue!15}  & \textbf{Base} & \textbf{RLVR} & \textbf{SRR} & \textbf{NDR} & \textbf{SDS} & \textbf{NSCR} & {\textbf{P}} & \textcolor{MyGreen}{\textbf{E}} & \textcolor{red}{\textbf{S}} & \textcolor{gray}{\textbf{O}} \\ \midrule
\rowcolor{purple!10} \multicolumn{11}{c}{\textbf{\textit{Math Reasoning Benchmarks (pass@8192)}}} \\
\midrule \rowcolor{yellow!8} AIME2024 & 93.3\% & 83.3\% & 0.893 & 0.000 & 0.000 & -0.107 & {\textbf{25}} & \textcolor{MyGreen}{\textbf{0}} & \textcolor{red}{\textbf{3}} & \textcolor{gray}{\textbf{2}} \\
\rowcolor{yellow!5}
AIME2025 & 80.0\% & 73.3\% & 0.833 & 0.091 & 0.164 & -0.077 & {\textbf{20}} & \textcolor{MyGreen}{\textbf{2}} & \textcolor{red}{\textbf{4}} & \textcolor{gray}{\textbf{4}} \\
\rowcolor{yellow!8}
AMC23 & 100.0\% & 100.0\% & 1.000 & 0.000 & 0.000 & 0.000 & {\textbf{40}} & \textcolor{MyGreen}{\textbf{0}} & \textcolor{red}{\textbf{0}} & \textcolor{gray}{\textbf{0}} \\
\rowcolor{yellow!5}
MATH500 & 99.6\% & 99.4\% & 0.998 & 0.000 & 0.000 & -0.002 & {\textbf{497}} & \textcolor{MyGreen}{\textbf{0}} & \textcolor{red}{\textbf{1}} & \textcolor{gray}{\textbf{2}} \\
\rowcolor{yellow!8}
Minerva & 71.7\% & 63.6\% & 0.887 & 0.000 & 0.000 & -0.113 & {\textbf{173}} & \textcolor{MyGreen}{\textbf{0}} & \textcolor{red}{\textbf{22}} & \textcolor{gray}{\textbf{77}} \\
\rowcolor{yellow!5}
Olympiad & 92.7\% & 89.3\% & 0.958 & 0.005 & 0.010 & -0.037 & {\textbf{600}} & \textcolor{MyGreen}{\textbf{3}} & \textcolor{red}{\textbf{26}} & \textcolor{gray}{\textbf{46}} \\
\midrule
\rowcolor{teal!10}
\multicolumn{11}{c}{\textbf{\textit{Non-Math Reasoning Benchmarks (pass@2048)}}} \\
\midrule
\rowcolor{cyan!8}
SimpleQA & 23.3\% & 18.0\% & 0.743 & 0.038 & 0.073 & -0.221 & {\textbf{75}} & \textcolor{MyGreen}{\textbf{3}} & \textcolor{red}{\textbf{26}} & \textcolor{gray}{\textbf{329}} \\
\rowcolor{cyan!5}
LiveBench-R & 100.0\% & 94.0\% & 0.940 & 0.000 & 0.000 & -0.060 & {\textbf{94}} & \textcolor{MyGreen}{\textbf{0}} & \textcolor{red}{\textbf{6}} & \textcolor{gray}{\textbf{0}} \\
\rowcolor{cyan!8}
LiveBench-C & 62.5\% & 56.2\% & 0.838 & 0.069 & 0.128 & -0.094 & {\textbf{67}} & \textcolor{MyGreen}{\textbf{5}} & \textcolor{red}{\textbf{13}} & \textcolor{gray}{\textbf{43}} \\
\rowcolor{cyan!5}
LiveBench-L & 26.0\% & 24.0\% & 0.769 & 0.167 & 0.274 & -0.067 & {\textbf{10}} & \textcolor{MyGreen}{\textbf{2}} & \textcolor{red}{\textbf{3}} & \textcolor{gray}{\textbf{35}} \\
\rowcolor{cyan!8}
SciBench & 94.1\% & 90.5\% & 0.946 & 0.016 & 0.031 & -0.038 & {\textbf{616}} & \textcolor{MyGreen}{\textbf{10}} & \textcolor{red}{\textbf{35}} & \textcolor{gray}{\textbf{31}} \\
\rowcolor{cyan!8}
LiveCodeBench v5 & 46.4\% & 43.0\% & 0.860 & 0.072 & 0.133 & -0.069 & {\textbf{129}} & \textcolor{MyGreen}{\textbf{10}} & \textcolor{red}{\textbf{21}} & \textcolor{gray}{\textbf{163}} \\
\rowcolor{cyan!8}
LiveCodeBench v6 & 43.5\% & 42.0\% & 0.947 & 0.018 & 0.036 & -0.034 & {\textbf{54}} & \textcolor{MyGreen}{\textbf{1}} & \textcolor{red}{\textbf{3}} & \textcolor{gray}{\textbf{73}} \\
\midrule \midrule
\rowcolor{violet!15}
\multicolumn{11}{c}{\textbf{Aggregate Statistics}} \\
\midrule
\rowcolor{orange!12}
\textbf{Math} & -- & -- & 0.960 & 0.0037 & 0.0073 & -0.036 & {\textbf{1355}} & \textcolor{MyGreen}{\textbf{5}} & \textcolor{red}{\textbf{56}} & \textcolor{gray}{\textbf{131}} \\
\rowcolor{pink!12}
\textbf{Non-Math} & -- & -- & 0.907 & 0.0288 & 0.0558 & -0.064 & {\textbf{1045}} & \textcolor{MyGreen}{\textbf{31}} & \textcolor{red}{\textbf{107}} & \textcolor{gray}{\textbf{674}} \\
\rowcolor{black!15}
\textbf{Overall} & \textbf{--} & \textbf{--} & 0.936 & 0.0148 & 0.0291 & -0.049 & {\textbf{2400}} & \textcolor{MyGreen}{\textbf{36}} & \textcolor{red}{\textbf{163}} & \textcolor{gray}{\textbf{805}} \\ \bottomrule
\end{tabular}
\end{supporttable}
\end{table*}

\begin{table*}[t]
\centering
\caption{Support dynamics metrics and \texttt{pass@k} performance of AceReason-Nemotron-1-7B, compared with its base model, DeepSeek-7B. }
\label{tab:support_nemotron}
\begin{supporttable}
\begin{tabular}{@{}c|cc|cccc|cccc@{}} \toprule
\rowcolor{blue!15}
\multirow{2}{*}{\textbf{Dataset}} & \multicolumn{2}{c|}{\textbf{pass@k Performance}} & \multicolumn{4}{c|}{\textbf{Support Dynamics Metrics}} & \multicolumn{4}{c}{\textbf{Support Counts}} \\
\cmidrule(lr){2-3} \cmidrule(lr){4-7} \cmidrule(lr){8-11}
\rowcolor{blue!15}
 & \textbf{Base} & \textbf{RLVR} & \textbf{SRR} & \textbf{NDR} & \textbf{SDS} & \textbf{NSCR} & {\textbf{P}} & \textcolor{MyGreen}{\textbf{E}} & \textcolor{red}{\textbf{S}} & \textcolor{gray}{\textbf{O}} \\
\midrule
\rowcolor{purple!10}
\multicolumn{11}{c}{\textbf{\textit{Math Reasoning Benchmarks (pass@8192)}}} \\
\midrule
\rowcolor{yellow!8}
AIME2024 & 93.3\% & 93.3\% & 1.000 & 0.000 & 0.000 & 0.000 & {\textbf{28}} & \textcolor{MyGreen}{\textbf{0}} & \textcolor{red}{\textbf{0}} & \textcolor{gray}{\textbf{2}} \\
\rowcolor{yellow!5}
AIME2025 & 100.0\% & 100.0\% & 1.000 & 0.000 & 0.000 & 0.000 & {\textbf{30}} & \textcolor{MyGreen}{\textbf{0}} & \textcolor{red}{\textbf{0}} & \textcolor{gray}{\textbf{0}} \\
\rowcolor{yellow!8}
AMC23 & 100.0\% & 100.0\% & 1.000 & 0.000 & 0.000 & 0.000 & {\textbf{40}} & \textcolor{MyGreen}{\textbf{0}} & \textcolor{red}{\textbf{0}} & \textcolor{gray}{\textbf{0}} \\
\rowcolor{yellow!5}
MATH500 & 99.8\% & 99.8\% & 1.000 & 0.000 & 0.000 & 0.000 & {\textbf{499}} & \textcolor{MyGreen}{\textbf{0}} & \textcolor{red}{\textbf{0}} & \textcolor{gray}{\textbf{1}} \\
\rowcolor{yellow!8}
Minerva & 71.7\% & 71.0\% & 0.985 & 0.005 & 0.010 & -0.010 & {\textbf{192}} & \textcolor{MyGreen}{\textbf{1}} & \textcolor{red}{\textbf{3}} & \textcolor{gray}{\textbf{76}} \\
\rowcolor{yellow!5}
Olympiad & 96.0\% & 95.7\% & 0.991 & 0.006 & 0.012 & -0.003 & {\textbf{642}} & \textcolor{MyGreen}{\textbf{4}} & \textcolor{red}{\textbf{6}} & \textcolor{gray}{\textbf{23}} \\
\midrule
\rowcolor{teal!10}
\multicolumn{11}{c}{\textbf{\textit{Non-Math Reasoning Benchmarks (pass@2048)}}} \\
\midrule
\rowcolor{cyan!8}
SimpleQA & 38.6\% & 35.6\% & 0.862 & 0.065 & 0.121 & -0.073 & {\textbf{144}} & \textcolor{MyGreen}{\textbf{10}} & \textcolor{red}{\textbf{23}} & \textcolor{gray}{\textbf{256}} \\
\rowcolor{cyan!5}
LiveBench-R & 100.0\% & 99.0\% & 0.990 & 0.000 & 0.000 & -0.010 & {\textbf{99}} & \textcolor{MyGreen}{\textbf{0}} & \textcolor{red}{\textbf{1}} & \textcolor{gray}{\textbf{0}} \\
\rowcolor{cyan!8}
LiveBench-C & 85.9\% & 85.9\% & 0.991 & 0.009 & 0.018 & 0.000 & {\textbf{109}} & \textcolor{MyGreen}{\textbf{1}} & \textcolor{red}{\textbf{1}} & \textcolor{gray}{\textbf{17}} \\
\rowcolor{cyan!5}
LiveBench-L & 24.0\% & 24.0\% & 0.833 & 0.167 & 0.278 & 0.000 & {\textbf{10}} & \textcolor{MyGreen}{\textbf{2}} & \textcolor{red}{\textbf{2}} & \textcolor{gray}{\textbf{36}} \\
\rowcolor{cyan!8}
SciBench & 94.7\% & 93.5\% & 0.982 & 0.006 & 0.012 & -0.012 & {\textbf{643}} & \textcolor{MyGreen}{\textbf{4}} & \textcolor{red}{\textbf{12}} & \textcolor{gray}{\textbf{33}} \\
\rowcolor{cyan!8}
LiveCodeBench v5 & 62.8\% & 62.5\% & 0.970 & 0.025 & 0.048 & -0.005 & {\textbf{197}} & \textcolor{MyGreen}{\textbf{5}} & \textcolor{red}{\textbf{6}} & \textcolor{gray}{\textbf{115}} \\
\rowcolor{cyan!8}
LiveCodeBench v6 & 64.1\% & 63.4\% & 0.976 & 0.012 & 0.024 & -0.012 & {\textbf{82}} & \textcolor{MyGreen}{\textbf{1}} & \textcolor{red}{\textbf{2}} & \textcolor{gray}{\textbf{46}} \\
\midrule \midrule
\rowcolor{violet!15}
\multicolumn{11}{c}{\textbf{Aggregate Statistics}} \\
\midrule
\rowcolor{orange!12}
\textbf{Math} & -- & -- & 0.994 & 0.003 & 0.007 & -0.003 & {\textbf{1431}} & \textcolor{MyGreen}{\textbf{5}} & \textcolor{red}{\textbf{9}} & \textcolor{gray}{\textbf{102}} \\
\rowcolor{pink!12}
\textbf{Non-Math} & -- & -- & 0.965 & 0.018 & 0.035 & -0.018 & {\textbf{1284}} & \textcolor{MyGreen}{\textbf{23}} & \textcolor{red}{\textbf{47}} & \textcolor{gray}{\textbf{503}} \\
\rowcolor{black!15}
\textbf{Overall} & \textbf{--} & \textbf{--} & 0.980 & 0.010 & 0.020 & -0.010 & {\textbf{2715}} & \textcolor{MyGreen}{\textbf{28}} & \textcolor{red}{\textbf{56}} & \textcolor{gray}{\textbf{605}} \\ \bottomrule
\end{tabular}
\end{supporttable}
\end{table*}

\begin{table*}[t]
\centering
\caption{Support dynamics metrics and \texttt{pass@k} performance of Skywork-OR1-7B, compared with its base model, DeepSeek-R1-Distill-Qwen-7B.}
\label{tab:support_skywork}
\begin{supporttable}
\begin{tabular}{@{}c|cc|cccc|cccc@{}} \toprule
\rowcolor{blue!15}
\multirow{2}{*}{\textbf{Dataset}} & \multicolumn{2}{c|}{\textbf{pass@k Performance}} & \multicolumn{4}{c|}{\textbf{Support Dynamics Metrics}} & \multicolumn{4}{c}{\textbf{Support Counts}} \\
\cmidrule(lr){2-3} \cmidrule(lr){4-7} \cmidrule(lr){8-11}
\rowcolor{blue!15}
 & \textbf{Base} & \textbf{RLVR} & \textbf{SRR} & \textbf{NDR} & \textbf{SDS} & \textbf{NSCR} & {\textbf{P}} & \textcolor{MyGreen}{\textbf{E}} & \textcolor{red}{\textbf{S}} & \textcolor{gray}{\textbf{O}} \\
\midrule
\rowcolor{purple!10}
\multicolumn{11}{c}{\textbf{\textit{Math Reasoning Benchmarks (pass@8192)}}} \\
\midrule
\rowcolor{yellow!8}
AIME2024 & 93.3\% & 93.3\% & 1.000 & 0.000 & 0.000 & 0.000 & {\textbf{28}} & \textcolor{MyGreen}{\textbf{0}} & \textcolor{red}{\textbf{0}} & \textcolor{gray}{\textbf{2}} \\
\rowcolor{yellow!5}
AIME2025 & 100.0\% & 100.0\% & 1.000 & 0.000 & 0.000 & 0.000 & {\textbf{30}} & \textcolor{MyGreen}{\textbf{0}} & \textcolor{red}{\textbf{0}} & \textcolor{gray}{\textbf{0}} \\
\rowcolor{yellow!8}
AMC23 & 100.0\% & 100.0\% & 1.000 & 0.000 & 0.000 & 0.000 & {\textbf{40}} & \textcolor{MyGreen}{\textbf{0}} & \textcolor{red}{\textbf{0}} & \textcolor{gray}{\textbf{0}} \\
\rowcolor{yellow!5}
MATH500 & 99.8\% & 99.8\% & 1.000 & 0.000 & 0.000 & 0.000 & {\textbf{499}} & \textcolor{MyGreen}{\textbf{0}} & \textcolor{red}{\textbf{0}} & \textcolor{gray}{\textbf{1}} \\
\rowcolor{yellow!8}
Minerva & 71.7\% & 71.3\% & 0.985 & 0.010 & 0.020 & -0.005 & {\textbf{192}} & \textcolor{MyGreen}{\textbf{2}} & \textcolor{red}{\textbf{3}} & \textcolor{gray}{\textbf{75}} \\
\rowcolor{yellow!5}
Olympiad & 96.0\% & 91.4\% & 0.952 & 0.000 & 0.000 & -0.048 & {\textbf{617}} & \textcolor{MyGreen}{\textbf{0}} & \textcolor{red}{\textbf{31}} & \textcolor{gray}{\textbf{27}} \\
\midrule
\rowcolor{teal!10}
\multicolumn{11}{c}{\textbf{\textit{Non-Math Reasoning Benchmarks (pass@2048)}}} \\
\midrule
\rowcolor{cyan!8}
SimpleQA & 38.6\% & 37.0\% & 0.880 & 0.081 & 0.149 & -0.039 & {\textbf{147}} & \textcolor{MyGreen}{\textbf{13}} & \textcolor{red}{\textbf{20}} & \textcolor{gray}{\textbf{253}} \\
\rowcolor{cyan!5}
LiveBench-R & 100.0\% & 98.0\% & 0.980 & 0.000 & 0.000 & -0.020 & {\textbf{98}} & \textcolor{MyGreen}{\textbf{0}} & \textcolor{red}{\textbf{2}} & \textcolor{gray}{\textbf{0}} \\
\rowcolor{cyan!8}
LiveBench-C & 85.9\% & 85.2\% & 0.991 & 0.000 & 0.000 & -0.009 & {\textbf{109}} & \textcolor{MyGreen}{\textbf{0}} & \textcolor{red}{\textbf{1}} & \textcolor{gray}{\textbf{18}} \\
\rowcolor{cyan!5}
LiveBench-L & 24.0\% & 22.0\% & 0.917 & 0.000 & 0.000 & -0.083 & {\textbf{11}} & \textcolor{MyGreen}{\textbf{0}} & \textcolor{red}{\textbf{1}} & \textcolor{gray}{\textbf{38}} \\
\rowcolor{cyan!8}
SciBench & 94.7\% & 92.8\% & 0.974 & 0.006 & 0.012 & -0.020 & {\textbf{638}} & \textcolor{MyGreen}{\textbf{4}} & \textcolor{red}{\textbf{17}} & \textcolor{gray}{\textbf{33}} \\
\rowcolor{cyan!8}
LiveCodeBench v5 & 62.8\% & 62.2\% & 0.966 & 0.025 & 0.049 & -0.010 & {\textbf{196}} & \textcolor{MyGreen}{\textbf{5}} & \textcolor{red}{\textbf{7}} & \textcolor{gray}{\textbf{115}} \\
\rowcolor{cyan!8}
LiveCodeBench v6 & 64.1\% & 62.6\% & 0.952 & 0.024 & 0.048 & -0.023 & {\textbf{80}} & \textcolor{MyGreen}{\textbf{2}} & \textcolor{red}{\textbf{4}} & \textcolor{gray}{\textbf{45}} \\
\midrule \midrule
\rowcolor{violet!15}
\multicolumn{11}{c}{\textbf{Aggregate Statistics}} \\
\midrule
\rowcolor{orange!12}
\textbf{Math} & -- & -- & 0.976 & 0.001 & 0.003 & -0.022 & {\textbf{1406}} & \textcolor{MyGreen}{\textbf{2}} & \textcolor{red}{\textbf{34}} & \textcolor{gray}{\textbf{105}} \\
\rowcolor{pink!12}
\textbf{Non-Math} & -- & -- & 0.961 & 0.018 & 0.036 & -0.021 & {\textbf{1279}} & \textcolor{MyGreen}{\textbf{24}} & \textcolor{red}{\textbf{52}} & \textcolor{gray}{\textbf{502}} \\
\rowcolor{black!15}
\textbf{Overall} & \textbf{--} & \textbf{--} & 0.969 & 0.010 & 0.019 & -0.021 & {\textbf{2685}} & \textcolor{MyGreen}{\textbf{26}} & \textcolor{red}{\textbf{86}} & \textcolor{gray}{\textbf{607}} \\ \bottomrule
\end{tabular}
\end{supporttable}
\end{table*}

\begin{table*}[t]
\centering
\caption{Support dynamics metrics and \texttt{pass@k} performance of Nemotron-1-14B, compared with its base model, DeepSeek-R1-Distill-Qwen-14B.}
\label{tab:support_nemo}
\begin{supporttable}
\begin{tabular}{@{}c|cc|cccc|cccc@{}} \toprule
\rowcolor{blue!15}
\multirow{2}{*}{\textbf{Dataset}} & \multicolumn{2}{c|}{\textbf{pass@k Performance}} & \multicolumn{4}{c|}{\textbf{Support Dynamics Metrics}} & \multicolumn{4}{c}{\textbf{Support Counts}} \\
\cmidrule(lr){2-3} \cmidrule(lr){4-7} \cmidrule(lr){8-11}
\rowcolor{blue!15}
 & \textbf{Base} & \textbf{RLVR} & \textbf{SRR} & \textbf{NDR} & \textbf{SDS} & \textbf{NSCR} & {\textbf{P}} & \textcolor{MyGreen}{\textbf{E}} & \textcolor{red}{\textbf{S}} & \textcolor{gray}{\textbf{O}} \\
\midrule
\rowcolor{purple!10}
\multicolumn{11}{c}{\textbf{\textit{Math Reasoning Benchmarks (pass@4096)}}} \\
\midrule
\rowcolor{yellow!8}
AIME2024 & 96.7\% & 93.3\% & 0.966 & 0.000 & 0.000 & -0.034 & {\textbf{28}} & \textcolor{MyGreen}{\textbf{0}} & \textcolor{red}{\textbf{1}} & \textcolor{gray}{\textbf{1}} \\
\rowcolor{yellow!5}
AIME2025 & 100.0\% & 96.7\% & 0.967 & 0.000 & 0.000 & -0.033 & {\textbf{29}} & \textcolor{MyGreen}{\textbf{0}} & \textcolor{red}{\textbf{1}} & \textcolor{gray}{\textbf{0}} \\
\rowcolor{yellow!8}
AMC23 & 100.0\% & 100.0\% & 1.000 & 0.000 & 0.000 & 0.000 & {\textbf{40}} & \textcolor{MyGreen}{\textbf{0}} & \textcolor{red}{\textbf{0}} & \textcolor{gray}{\textbf{0}} \\
\rowcolor{yellow!5}
MATH500 & 99.8\% & 99.8\% & 1.000 & 0.000 & 0.000 & 0.000 & {\textbf{499}} & \textcolor{MyGreen}{\textbf{0}} & \textcolor{red}{\textbf{0}} & \textcolor{gray}{\textbf{1}} \\
\rowcolor{yellow!8}
Minerva & 71.7\% & 69.5\% & 0.959 & 0.011 & 0.021 & -0.030 & {\textbf{187}} & \textcolor{MyGreen}{\textbf{2}} & \textcolor{red}{\textbf{8}} & \textcolor{gray}{\textbf{75}} \\
\rowcolor{yellow!5}
Olympiad & 95.9\% & 95.6\% & 0.992 & 0.005 & 0.009 & -0.003 & {\textbf{642}} & \textcolor{MyGreen}{\textbf{3}} & \textcolor{red}{\textbf{5}} & \textcolor{gray}{\textbf{25}} \\
\midrule
\rowcolor{teal!10}
\multicolumn{11}{c}{\textbf{\textit{Non-Math Reasoning Benchmarks (pass@1024)}}} \\
\midrule
\rowcolor{cyan!8}
SimpleQA & 27.0\% & 26.8\% & 0.983 & 0.009 & 0.017 & -0.008 & {\textbf{115}} & \textcolor{MyGreen}{\textbf{1}} & \textcolor{red}{\textbf{2}} & \textcolor{gray}{\textbf{315}} \\
\rowcolor{cyan!5}
LiveBench-R & 99.0\% & 99.0\% & 1.000 & 0.000 & 0.000 & 0.000 & {\textbf{99}} & \textcolor{MyGreen}{\textbf{0}} & \textcolor{red}{\textbf{0}} & \textcolor{gray}{\textbf{1}} \\
\rowcolor{cyan!8}
LiveBench-C & 92.2\% & 92.2\% & 1.000 & 0.000 & 0.000 & 0.000 & {\textbf{118}} & \textcolor{MyGreen}{\textbf{0}} & \textcolor{red}{\textbf{0}} & \textcolor{gray}{\textbf{10}} \\
\rowcolor{cyan!5}
LiveBench-L & 46.0\% & 44.0\% & 0.957 & 0.000 & 0.000 & -0.043 & {\textbf{22}} & \textcolor{MyGreen}{\textbf{0}} & \textcolor{red}{\textbf{1}} & \textcolor{gray}{\textbf{27}} \\
\rowcolor{cyan!8}
SciBench & 93.1\% & 92.6\% & 0.992 & 0.003 & 0.006 & -0.005 & {\textbf{639}} & \textcolor{MyGreen}{\textbf{2}} & \textcolor{red}{\textbf{5}} & \textcolor{gray}{\textbf{46}} \\
\midrule \midrule
\rowcolor{violet!15}
\multicolumn{11}{c}{\textbf{Aggregate Statistics}} \\
\midrule
\rowcolor{orange!12}
\textbf{Math} & -- & -- & 0.990 & 0.0035 & 0.0070 & -0.0069 & {\textbf{1425}} & \textcolor{MyGreen}{\textbf{5}} & \textcolor{red}{\textbf{15}} & \textcolor{gray}{\textbf{102}} \\
\rowcolor{pink!12}
\textbf{Non-Math} & -- & -- & 0.992 & 0.0030 & 0.0060 & -0.0050 & {\textbf{993}} & \textcolor{MyGreen}{\textbf{3}} & \textcolor{red}{\textbf{8}} & \textcolor{gray}{\textbf{399}} \\
\rowcolor{black!15}
\textbf{Overall} & \textbf{--} & \textbf{--} & 0.991 & 0.0033 & 0.0066 & -0.0061 & {\textbf{2418}} & \textcolor{MyGreen}{\textbf{8}} & \textcolor{red}{\textbf{23}} & \textcolor{gray}{\textbf{501}} \\ \bottomrule
\end{tabular}
\end{supporttable}
\end{table*}

\begin{table*}[t]
\centering
\caption{Support dynamics metrics and \texttt{pass@k} performance of Phi4-Reason-Plus-14B, compared with its base model -- Phi4-Reason-14B. }
\label{tab:support_phi4}
\begin{supporttable}
\begin{tabular}{@{}c|cc|cccc|cccc@{}} \toprule
\rowcolor{blue!15}
\multirow{2}{*}{\textbf{Dataset}} & \multicolumn{2}{c|}{\textbf{pass@k Performance}} & \multicolumn{4}{c|}{\textbf{Support Dynamics Metrics}} & \multicolumn{4}{c}{\textbf{Support Counts}} \\
\cmidrule(lr){2-3} \cmidrule(lr){4-7} \cmidrule(lr){8-11}
\rowcolor{blue!15}
 & \textbf{Base} & \textbf{RLVR} & \textbf{SRR} & \textbf{NDR} & \textbf{SDS} & \textbf{NSCR} & {\textbf{P}} & \textcolor{MyGreen}{\textbf{E}} & \textcolor{red}{\textbf{S}} & \textcolor{gray}{\textbf{O}} \\
\midrule
\rowcolor{purple!10}
\multicolumn{11}{c}{\textbf{\textit{Math Reasoning Benchmarks (pass@4096)}}} \\
\midrule
\rowcolor{yellow!8}
AIME2024 & 96.7\% & 96.7\% & 1.000 & 0.000 & 0.000 & 0.000 & {\textbf{29}} & \textcolor{MyGreen}{\textbf{0}} & \textcolor{red}{\textbf{0}} & \textcolor{gray}{\textbf{1}} \\
\rowcolor{yellow!5}
AIME2025 & 100.0\% & 100.0\% & 1.000 & 0.000 & 0.000 & 0.000 & {\textbf{30}} & \textcolor{MyGreen}{\textbf{0}} & \textcolor{red}{\textbf{0}} & \textcolor{gray}{\textbf{0}} \\
\rowcolor{yellow!8}
AMC23 & 100.0\% & 100.0\% & 1.000 & 0.000 & 0.000 & 0.000 & {\textbf{40}} & \textcolor{MyGreen}{\textbf{0}} & \textcolor{red}{\textbf{0}} & \textcolor{gray}{\textbf{0}} \\
\rowcolor{yellow!5}
MATH500 & 100.0\% & 99.8\% & 0.998 & 0.000 & 0.000 & -0.002 & {\textbf{499}} & \textcolor{MyGreen}{\textbf{0}} & \textcolor{red}{\textbf{1}} & \textcolor{gray}{\textbf{0}} \\
\rowcolor{yellow!8}
Minerva & 66.2\% & 65.4\% & 0.972 & 0.017 & 0.033 & -0.011 & {\textbf{175}} & \textcolor{MyGreen}{\textbf{3}} & \textcolor{red}{\textbf{5}} & \textcolor{gray}{\textbf{89}} \\
\rowcolor{yellow!5}
Olympiad & 94.8\% & 94.7\% & 0.991 & 0.008 & 0.016 & -0.002 & {\textbf{634}} & \textcolor{MyGreen}{\textbf{5}} & \textcolor{red}{\textbf{6}} & \textcolor{gray}{\textbf{30}} \\
\midrule
\rowcolor{teal!10}
\multicolumn{11}{c}{\textbf{\textit{Non-Math Reasoning Benchmarks (pass@1024)}}} \\
\midrule
\rowcolor{cyan!8}
SimpleQA & 37.9\% & 37.4\% & 0.970 & 0.019 & 0.036 & -0.012 & {\textbf{159}} & \textcolor{MyGreen}{\textbf{3}} & \textcolor{red}{\textbf{5}} & \textcolor{gray}{\textbf{266}} \\
\rowcolor{cyan!5}
LiveBench-R & 100.0\% & 100.0\% & 1.000 & 0.000 & 0.000 & 0.000 & {\textbf{100}} & \textcolor{MyGreen}{\textbf{0}} & \textcolor{red}{\textbf{0}} & \textcolor{gray}{\textbf{0}} \\
\rowcolor{cyan!8}
LiveBench-C & 97.7\% & 96.9\% & 0.992 & 0.000 & 0.000 & -0.008 & {\textbf{124}} & \textcolor{MyGreen}{\textbf{0}} & \textcolor{red}{\textbf{1}} & \textcolor{gray}{\textbf{3}} \\
\rowcolor{cyan!5}
LiveBench-L & 74.0\% & 74.0\% & 1.000 & 0.000 & 0.000 & 0.000 & {\textbf{37}} & \textcolor{MyGreen}{\textbf{0}} & \textcolor{red}{\textbf{0}} & \textcolor{gray}{\textbf{13}} \\
\rowcolor{cyan!8}
SciBench & 94.2\% & 94.2\% & 0.992 & 0.008 & 0.015 & 0.000 & {\textbf{647}} & \textcolor{MyGreen}{\textbf{5}} & \textcolor{red}{\textbf{5}} & \textcolor{gray}{\textbf{35}} \\
\midrule \midrule
\rowcolor{violet!15}
\multicolumn{11}{c}{\textbf{Aggregate Statistics}} \\
\midrule
\rowcolor{orange!12}
\textbf{Math} & -- & -- & 0.992 & 0.0057 & 0.0112 & -0.0028 & {\textbf{1407}} & \textcolor{MyGreen}{\textbf{8}} & \textcolor{red}{\textbf{12}} & \textcolor{gray}{\textbf{120}} \\
\rowcolor{pink!12}
\textbf{Non-Math} & -- & -- & 0.990 & 0.0074 & 0.0148 & -0.0028 & {\textbf{1067}} & \textcolor{MyGreen}{\textbf{8}} & \textcolor{red}{\textbf{11}} & \textcolor{gray}{\textbf{317}} \\
\rowcolor{black!15}
\textbf{Overall} & \textbf{--} & \textbf{--} & 0.991 & 0.0064 & 0.0128 & -0.0028 & {\textbf{2474}} & \textcolor{MyGreen}{\textbf{16}} & \textcolor{red}{\textbf{23}} & \textcolor{gray}{\textbf{437}} \\ \bottomrule
\end{tabular}
\end{supporttable}
\end{table*}

\begin{table*}[t]
\centering
\caption{Support dynamics metrics and \texttt{pass@k} performance of vision-language model OVR-7B-RL, compared with its base model, OVR-7B-ColdStart, across visual math reasoning benchmarks. }
\label{tab:vlm_support_metrics}
\begin{supporttable}
\begin{tabular}{@{}c|cc|cccc|cccc@{}} \toprule
\rowcolor{blue!15}
\multirow{2}{*}{\textbf{Dataset}} & \multicolumn{2}{c|}{\textbf{pass@k Performance}} & \multicolumn{4}{c|}{\textbf{Support Dynamics Metrics}} & \multicolumn{4}{c}{\textbf{Support Counts}} \\
\cmidrule(lr){2-3} \cmidrule(lr){4-7} \cmidrule(lr){8-11}
\rowcolor{blue!15}
 & \textbf{Base} & \textbf{RLVR} & \textbf{SRR} & \textbf{NDR} & \textbf{SDS} & \textbf{NSCR} & {\textbf{P}} & \textcolor{MyGreen}{\textbf{E}} & \textcolor{red}{\textbf{S}} & \textcolor{gray}{\textbf{O}} \\
\midrule
\rowcolor{purple!10}
\multicolumn{11}{c}{\textbf{\textit{Math Reasoning Benchmarks (pass@8192)}}} \\
\midrule
\rowcolor{yellow!8}
MathVista & 49.1\% & 49.1\% & 0.998 & 0.002 & 0.004 & 0.000 & {\textbf{490}} & \textcolor{MyGreen}{\textbf{1}} & \textcolor{red}{\textbf{1}} & \textcolor{gray}{\textbf{508}} \\
\rowcolor{yellow!5}
MathVision & 96.7\% & 96.4\% & 0.990 & 0.007 & 0.014 & -0.003 & {\textbf{291}} & \textcolor{MyGreen}{\textbf{2}} & \textcolor{red}{\textbf{3}} & \textcolor{gray}{\textbf{8}} \\
\midrule \midrule
\rowcolor{violet!15}
\multicolumn{11}{c}{\textbf{Aggregate Statistics}} \\
\midrule
\rowcolor{orange!12}
\textbf{Math} & -- & -- & 0.995 & 0.0038 & 0.0076 & -0.0013 & {\textbf{781}} & \textcolor{MyGreen}{\textbf{3}} & \textcolor{red}{\textbf{4}} & \textcolor{gray}{\textbf{516}} \\  \bottomrule
\end{tabular}
\end{supporttable}
\end{table*}

\begin{table*}[t]
\centering
{\caption{Support dynamics metrics and \texttt{pass@256} performance of OLMo-2-0425-1B-DPO vs RLVR.}}
\label{tab:support_olmo}
\begin{supporttable}
\begin{tabular}{@{}c|cc|cccc|cccc@{}} \toprule \rowcolor{blue!15}
\multirow{2}{*}{\textbf{Dataset}}  & \multicolumn{2}{c|}{\textbf{pass@k Performance}}  & \multicolumn{4}{c|}{\textbf{Support Dynamics Metrics}} & \multicolumn{4}{c}{\textbf{Support Counts}} \\
\cmidrule(lr){2-3} \cmidrule(lr){4-7} \cmidrule(lr){8-11} \rowcolor{blue!15}
& \textbf{Base} & \textbf{RLVR} 
& \textbf{SRR} & \textbf{NDR} & \textbf{SDS} & \textbf{NSCR}
& {\textbf{P}} & \textcolor{MyGreen}{\textbf{E}} 
& \textcolor{red}{\textbf{S}} & \textcolor{gray}{\textbf{O}} \\ \midrule
\rowcolor{purple!10} \multicolumn{11}{c}{\textbf{\textit{Math Reasoning Benchmarks (pass@256)}}} \\ \midrule
\rowcolor{yellow!8}
AIME2024  & 20.00\% & 16.67\%  & 0.833 & 0.000 & 0.000 & -0.167  & {5} & \textcolor{MyGreen}{0} & \textcolor{red}{1} & \textcolor{gray}{24} \\
\rowcolor{yellow!5}
AIME2025  & 23.33\% & 20.00\%  & 0.571 & 0.333 & 0.421 & -0.111  & {4} & \textcolor{MyGreen}{2} & \textcolor{red}{3} & \textcolor{gray}{21} \\
\rowcolor{yellow!8} AMC23  & 85.00\% & 77.50\% 
& 0.912 & 0.000 & 0.000 & -0.088 
& {31} & \textcolor{MyGreen}{0} & \textcolor{red}{3} & \textcolor{gray}{6} \\
\rowcolor{yellow!5}
MATH500 
& 77.80\% & 76.20\% 
& 0.920 & 0.060 & 0.113 & -0.019 
& {358} & \textcolor{MyGreen}{23} & \textcolor{red}{31} & \textcolor{gray}{88} \\
\rowcolor{yellow!8}
Minerva-Math 
& 33.82\% & 32.35\% 
& 0.837 & 0.125 & 0.218 & -0.039 
& {77} & \textcolor{MyGreen}{11} & \textcolor{red}{15} & \textcolor{gray}{169} \\
\rowcolor{yellow!5}
Olympiad  & 49.93\% & 49.33\%  & 0.849 & 0.141 & 0.242 & -0.010  & {286} & \textcolor{MyGreen}{47} & \textcolor{red}{51} & \textcolor{gray}{291} \\
\midrule\midrule
\rowcolor{violet!15}
\multicolumn{11}{c}{\textbf{Aggregate Statistics}} \\\midrule
\rowcolor{orange!12} \textbf{Math}  & -- & --  & 0.880 & 0.098 & 0.177 &-0.022  & {761} & \textcolor{MyGreen}{83} & \textcolor{red}{104} & \textcolor{gray}{599} \\
\bottomrule
\end{tabular}
\end{supporttable}
\end{table*}

\section{Additional Results}

Table~\ref{tab:entropy} reports \texttt{avg@32} accuracy, response length, token-level entropy, and answer-level entropy across the math benchmarks. Token-entropy increases occur with opposite changes in response length: ProRL shortens its responses from an average of $10{,}422$ to $6{,}479$ tokens, whereas DAPO-32B lengthens its responses from $1{,}426$ to $4{,}989$ tokens.
\begin{table*}[t]
\centering
\caption{Accuracy (\texttt{avg@32}, \%), response length (tokens), and entropy across five math benchmarks. The Avg. column is the unweighted mean over benchmarks. \textnormal{Row colors: \colorbox{basegray}{\phantom{xx}} reference models, \colorbox{rlvrblue}{\phantom{xx}} RLVR models.} Relative to the corresponding reference checkpoint, RLVR increases accuracy and reduces answer entropy in every model--benchmark comparison; token entropy increases for ProRL and DAPO-32B and decreases for AceReason and Skywork.}
\label{tab:entropy}
\resizebox{\textwidth}{!}{%
\begin{tabular}{c|c|cccccc}
\toprule
\textbf{Metric} & \textbf{Model} & \textbf{AIME24} & \textbf{AMC23} & \textbf{MATH500} & \textbf{Minerva} & \textbf{Olympiad} & \textbf{Avg.} \\
\midrule 
  & \cellcolor{basegray} DeepSeek-1.5B & 31.15 & 72.81 & 85.01 & 32.18 & 51.55 & 54.54 \\
  & \cellcolor{rlvrblue} ProRL-1.5B        & 45.62 & 85.70 & 92.01 & 39.27 & 64.56 & 65.43 \\ \cline{2-8}\addlinespace[3pt]
  & \cellcolor{basegray} DeepSeek-7B       & 53.23 & 89.30 & 93.95 & 43.07 & 66.67 & 69.24 \\
  \texttt{avg@32}  & \cellcolor{rlvrblue}AceReason-7B      & 65.83 & 95.08 & 95.81 & 45.35 & 73.92 & 75.20 \\
  Acc. (\%) & \cellcolor{rlvrblue}Skywork-OR1-7B        &  67.40  &  93.59  & 95.73 & 43.81 & 73.05 & 74.71 \\ \cline{2-8}\addlinespace[3pt]
  & \cellcolor{basegray} DeepSeek-14B       & 67.81 & 95.39 & 95.28 & 46.43 & 72.06 & 75.39 \\
  & \cellcolor{rlvrblue}AceReason-14B      & 77.29 & 98.67 & 97.01 & 47.20 & 77.74 & 79.58 \\ \cline{2-8}\addlinespace[3pt]
  & \cellcolor{basegray} Qwen2.5-32B       & 18.12 & 55.23 & 75.84 & 24.55 & 41.40 & 43.03 \\
  & \cellcolor{rlvrblue}DAPO-32B          & 51.25 & 92.81 & 80.75 & 32.50 & 49.15 & 61.29 \\ \midrule
  & \cellcolor{basegray} DeepSeek-1.5B     & 16363 & 9979  & 5700  & 8194  & 11873 & 10422 \\
  & \cellcolor{rlvrblue}ProRL-1.5B        & 7786  & 6294  & 5070  & 6569  & 6678  & 6479 \\ \cline{2-8}\addlinespace[3pt]
  & \cellcolor{basegray} DeepSeek-7B       & 13613 & 6402  & 4125  & 5595  & 8988  & 7745 \\
  Response & \cellcolor{rlvrblue}AceReason-7B & 10740 & 5961  & 4313  & 6261  & 7703  & 6995 \\
  Length & \cellcolor{rlvrblue}Skywork-OR1-7B  & 15628 & 8282 & 5735 & 8742 & 12094 & 10096 \\ \cline{2-8}\addlinespace[3pt]
  & \cellcolor{basegray} DeepSeek-14B       & 11295 & 5735  & 3781  & 4919  & 8042  & 6755 \\
  & \cellcolor{rlvrblue}AceReason-14B      & 13871 & 7239  & 4622  & 7720  & 10033 & 8697 \\ \cline{2-8}\addlinespace[3pt]
  & \cellcolor{basegray} Qwen2.5-32B       & 1247  & 874   & 585   & 3544  & 881   & 1426 \\
  & \cellcolor{rlvrblue}DAPO-32B       & 6908  & 3157  & 3386  & 5665  & 5827  & 4989 \\
\midrule
  &\cellcolor{basegray}  DeepSeek-1.5B     & 0.45  & 0.40  & 0.42  & 0.49  & 0.44  & 0.44 \\
  & \cellcolor{rlvrblue}ProRL-1.5B        & 0.47$\blacktriangle$ & 0.51$\blacktriangle$ & 0.54$\blacktriangle$ & 0.55$\blacktriangle$ & 0.52$\blacktriangle$ & 0.52$\blacktriangle$ \\ \cline{2-8}\addlinespace[3pt]
  & \cellcolor{basegray} DeepSeek-7B       & 0.38  & 0.34  & 0.35  & 0.39  & 0.38  & 0.37 \\
  Token-Level & \cellcolor{rlvrblue} AceReason-7B & 0.18$\blacktriangledown$ & 0.23$\blacktriangledown$ & 0.27$\blacktriangledown$ & 0.24$\blacktriangledown$ & 0.23$\blacktriangledown$ & 0.23$\blacktriangledown$ \\
  Entropy & \cellcolor{rlvrblue} Skywork-OR1-7B & 0.14$\blacktriangledown$ & 0.16$\blacktriangledown$ & 0.19$\blacktriangledown$ & 0.17$\blacktriangledown$ & 0.16$\blacktriangledown$ & 0.16$\blacktriangledown$ \\ \cline{2-8}\addlinespace[3pt]
  & \cellcolor{basegray} DeepSeek-14B       & 0.33  & 0.30  & 0.32  & 0.35  & 0.33  & 0.33 \\
  & \cellcolor{rlvrblue}AceReason-14B      & 0.12$\blacktriangledown$ & 0.13$\blacktriangledown$ & 0.15$\blacktriangledown$ & 0.15$\blacktriangledown$ & 0.14$\blacktriangledown$ & 0.14$\blacktriangledown$ \\ \cline{2-8}\addlinespace[3pt]
  & \cellcolor{basegray} Qwen2.5-32B       & 0.17  & 0.16  & 0.15  & 0.28  & 0.15  & 0.18 \\
  & \cellcolor{rlvrblue} DAPO-32B          & 0.26$\blacktriangle$ & 0.19$\blacktriangle$ & 0.27$\blacktriangle$ & 0.44$\blacktriangle$ & 0.30$\blacktriangle$ & 0.29$\blacktriangle$ \\
\midrule
  & \cellcolor{basegray} DeepSeek-1.5B     & 2.15  & 0.91  & 0.46  & 1.65  & 1.33  & 1.30 \\
  & \cellcolor{rlvrblue}ProRL-1.5B        & 1.24  & 0.35  & 0.18  & 0.90  & 0.63  & 0.66 \\ \cline{2-8}\addlinespace[3pt]
  & \cellcolor{basegray} DeepSeek-7B       & 1.47  & 0.36  & 0.18  & 0.96  & 0.80  & 0.75 \\
  Answer-Level & \cellcolor{rlvrblue} AceReason-7B & 0.96  & 0.14  & 0.11  & 0.77  & 0.53  & 0.50 \\
  Entropy & \cellcolor{rlvrblue} Skywork-OR1-7B  & 0.97 & 0.20 & 0.12 & 0.80 & 0.58 & 0.54 \\ \cline{2-8}\addlinespace[3pt]
  & \cellcolor{basegray} DeepSeek-14B       & 1.01  & 0.14  & 0.13  & 0.83  & 0.59  & 0.54 \\
  & \cellcolor{rlvrblue} AceReason-14B      & 0.66  & 0.06  & 0.07  & 0.67  & 0.44  & 0.38 \\ \cline{2-8}\addlinespace[3pt]
  & \cellcolor{basegray} Qwen2.5-32B       & 2.37  & 1.32  & 0.68  & 2.27  & 1.41  & 1.61 \\
  & \cellcolor{rlvrblue} DAPO-32B          & 1.12  & 0.09  & 0.26  & 0.96  & 0.63  & 0.61 \\
\bottomrule
\end{tabular}}
\end{table*}

\section{Experimental Details}
\label{app:exp}
We detail the datasets and decoding settings, then document the prompt and answer-processing changes used to evaluate base and ProRL responses on Reasoning Gym.

\subsection{Evaluation Settings}
We employed vLLM~\citep{vllm} as the inference backend. For all models, we utilized a sampling temperature of $0.6$, a $top\_p$ value of $0.95$, and a maximum response length of $32768$ tokens.

\subsection{Datasets}
\paragraph{Math benchmarks.}
We utilized the complete datasets from MATH500~\citep{hendrycks2021measuring}, Minerva~\citep{lewkowycz2022solving}, OlympiadBench~\citep{he2024olympiadbench}, AIME 2024, AIME 2025, and AMC 2023 for evaluating LLMs. For vision-language models, we evaluated on the testmini sets of MathVision~\citep{mathvision} and MathVista~\citep{mathvista}.

\paragraph{Non-math benchmarks.}
For SimpleQA~\citep{wei2024measuring}, we uniformly sampled 10\% of the original dataset (433 samples) to enable efficient large-scale evaluation at large sampling budgets.
For LiveBench~\citep{white2025livebench}, we used the 2024-11-25 version available on HuggingFace. Because the support analysis requires a binary verification result, we selected tasks with correct/incorrect judgments and excluded tasks scored with intermediate floating-point values. Based on this selection criterion, we evaluated the following subsets: $web\_of\_lies\_v2$ and $spatial$ subsets for Reasoning tasks (LiveBench-R), the $typos$ subset for Language tasks (LiveBench-L), and all available data for Coding tasks (LiveBench-C). For SciBench~\citep{wang2023scibench}, we evaluated on the complete dataset. For LiveCodeBench~\citep{jain2024livecodebench}, we evaluated the dataset on both v5 and v6 versions. Due to computational efficiency considerations, we conducted LiveCodeBench evaluation exclusively on 1.5B and 7B models, excluding the 14B variants from this particular benchmark.

\paragraph{Reasoning Gym.}
For Reasoning Gym~\citep{stojanovski2025reasoninggymreasoningenvironments}, we employ the \texttt{easy} set from the version updated after commit \textit{17a8431} in its repository as our \texttt{default} task configuration. This choice ensures consistency with the \texttt{default} task configuration used in prior evaluations, maintaining comparable experimental conditions. Additionally, we use the \texttt{hard} set as our challenging evaluation benchmark.

\subsection{Answer Processing Enhancement in Reasoning Gym}
\label{app:reasonining_gym_evaluation}

Evaluating the base model on Reasoning Gym revealed frequent mismatches between its answer format and the verifier's expected input. ProRL, which was trained on Reasoning Gym data, generally followed the expected format. We therefore clarified the format instructions in the prompts and expanded answer extraction to recognize alternative representations. These changes distinguish our evaluation protocol from the one used to report ProRL's original Reasoning Gym scores.

\subsubsection{General Answer Extraction Protocol}

First, we enhanced the answer extraction protocol with a hierarchical, priority-based extraction mechanism that processes responses through multiple fallback levels. Each level attempts to capture the model's intended answer, and successful extraction at any level bypasses later extraction fallbacks.

The strategy first attempts to extract content using the Reasoning Gym's \texttt{extract\_answer()} function, which captures answers within \texttt{<answer></answer>} tags. This approach is given the highest priority because these tags are Reasoning Gym's default format. When this method fails, the system searches for content within the final \texttt{\textbackslash boxed\{\}} formatting.

For dice tasks using the base model, failed \texttt{extract\_answer()} attempts trigger additional processing through Lighteval~\citep{lighteval}'s \texttt{math\_normalizer()} function. This function handles \texttt{\textbackslash boxed\{\}} capture and converts $a/b$ fractions to \LaTeX{} format \texttt{\textbackslash frac\{a\}\{b\}}. When \texttt{extract\_answer()} successfully captures $a/b$ fraction answers, the system applies Lighteval's \texttt{fix\_a\_slash\_b()} function to achieve the same \LaTeX{} conversion.

For non-dice tasks or when using ProRL models, failed \texttt{extract\_answer()} attempts utilize Lighteval's \texttt{last\_boxed\_only\_string()} and \texttt{remove\_boxed()} functions. These functions locate content within the final \texttt{\textbackslash boxed\{\}}, primarily addressing cases where base model prompt modifications shifted from answer tags to boxed formatting.

As a final fallback, the system extracts content following \texttt{</think>} tags when all previous methods fail, and the response contains these markers. This safety mechanism captures base model responses that ignore formatting requirements in lengthy tasks.

\subsubsection{Task-Specific Processing Modifications}

Our core answer-processing pipeline applies to both models, with additional steps designed primarily to address format compatibility issues commonly encountered in base-model responses.
Specifically, the processing logic for each task is enhanced as follows:

\paragraph{$dice$}
The ground truth for dice tasks uses $a/b$ fraction format. Base models frequently express fractions in \LaTeX{} format, requiring format standardization for accurate evaluation. For base models only, we convert ground truth fractions from $a/b$ format to \LaTeX{} format \texttt{\textbackslash frac\{a\}\{b\}} to ensure both model answers and ground truth use consistent \LaTeX{} formatting. ProRL dice processing maintains $a/b$ formatting for both model answers and ground truth.

\paragraph{$prime\_factorization$}
The ground truth format requires answers to be combinations of numbers and multiplication symbol (i.e., $\times$) only. We implement three key modifications to ensure compatibility with this requirement. First, we standardize \LaTeX{} multiplication symbols by replacing \texttt{\textbackslash times} with $\times$ to meet the evaluation requirements, as base models frequently use \LaTeX{} multiplication symbols instead of standard multiplication signs. Second, we expand \LaTeX{} exponentiation by converting formats like \texttt{a\textasciicircum b} into repeated multiplication ($a \times a \times \ldots \times a$ for $b$ iterations), preventing errors when base models consolidate repeated factors into exponential notation. Third, we process response equations by retaining only right-side content when answers contain equals signs, transforming responses like ``561 = 3 $\times$ 11 $\times$ 17'' to ``3 $\times$ 11 $\times$ 17'' to eliminate question restatement that base models commonly include.

\paragraph{$palindrome\_generation$}
The ground-truth format expects palindromic character strings (sequences that read the same forward and backward). We remove excess whitespace to address frequent spacing issues in base model responses. This transformation converts spaced responses like ``k h g a g h k'' to ``khgaghk'', so that spacing does not affect evaluation of the character sequence.

\paragraph{$advanced\_geometry$}
The ground truth format requires floating-point numbers. Our processing includes three main steps to address \LaTeX{} formatting issues commonly encountered in base-model output. First, we remove redundant \LaTeX{} expressions by eliminating \texttt{\textbackslash left} and \texttt{\textbackslash right} markers while converting \texttt{\textasciicircum\textbackslash circ} to ${}^{\circ}$ symbol, addressing base models' tendency to use \LaTeX{} for brackets and degree symbols. Second, we convert \LaTeX{} numerical expressions, transforming fractions \texttt{\textbackslash frac\{a\}\{b\}} and other \LaTeX{} formats (\texttt{\textbackslash sqrt\{a\}}, \texttt{\textbackslash sin\{a\}}, \texttt{\textbackslash log\{a\}}, \texttt{\textbackslash pi}, etc.) into three-decimal floating-point numbers using the \texttt{latex2sympy2\_extended} library's \texttt{latex2sympy()} function. Third, we evaluate arithmetic expressions containing radicals (such as $2\sqrt{16} + 5\sqrt{4} - 3$) by converting them into three-decimal floating-point numbers using Python's built-in mathematical functions, handling cases where base models output final results as arithmetic expressions rather than computed values.

\paragraph{$power\_function$}
The ground-truth format uses scientific notation. We convert mixed \LaTeX{} and arithmetic symbol scientific notation to ensure format consistency. The system transforms patterns like ``$-2.36 \times 10^{-16}$'' or ``$1.5 \times 10^{5}$'' to e-notation format (``-2.36e-16'', ``1.5e5''), preventing numerically correct but format-incompatible evaluation errors when base models use mixed \LaTeX{} and arithmetic symbols for scientific notation.

\paragraph{$arc\_1d$}
The ground truth format requires space-separated digit sequences. We handle two types of responses to meet this grid format requirement. For pure numerical responses, we insert spaces between consecutive digits, converting sequences like ``22220000000000000000111'' to ``2 2 2 2 0 0 0 0 0 0 0 0 0 0 0 0 0 0 0 0 1 1 1''. For mixed numerical and textual responses, we extract digits and insert spaces, transforming \LaTeX{} grid formats like \texttt{\textbackslash begin\{array\}\{cccc\} 0 \& 0 \& 0 \& 0 \& 0 \& 0 \& 0 \& 0 \& 7 \& 3 \& 0 \& 0 \& 4 \& 6 \textbackslash\textbackslash \textbackslash end\{array\}} to ``0 0 0 0 0 0 0 0 7 3 0 0 4 6'', addressing base models' tendency to output correct answers in \LaTeX{} grid format.

\paragraph{$boxnet$}
The ground truth format requires dictionary list formatting \texttt{[\{key: value\}, ...]}. We implement comprehensive cleaning of the JSON format to meet these evaluation requirements. Our processing pipeline includes several steps: rejecting pure numerical responses to prevent non-JSON format interference; removing JSON markdown wrappers that eliminate \texttt{```json \{content\} ```} markers; converting single dictionaries to single-element dictionary lists (\texttt{dict} $\rightarrow$ \texttt{[dict]}); and filtering illegal elements by removing non-dictionary components from JSON lists. Additionally, we clean nested structure values within individual dictionary entries. For nested lists, we extract the first element as the value (\texttt{[\{key1: [value1, value2, ...]\}, ...]} $\rightarrow$ \texttt{[\{key1: value1\}, ...]}). For nested dictionaries, we select matching key values when available (\texttt{[\{key1: \{key1: value1, key2: value2, ...\}\}, ...]} $\rightarrow$ \texttt{[\{key1: value1\}, ...]}) or default to the first element value when keys don't match (\texttt{[\{key1: \{key2: value2, key3: value3\}\}, ...]} $\rightarrow$ \texttt{[\{key1: value2\}, ...]}).

\subsection{Entropy Analysis}
\paragraph{Setup.} 
All models use a sampling temperature of $0.6$, a $top\_p$ value of $0.95$, and a maximum response length of $32768$ tokens. We generate 32 completions per question. For each question, accuracy and response length are averaged over the 32 completions; token entropy is averaged first over positions within each completion and then over completions. Answer entropy is computed from the frequencies of extracted final answers across the 32 completions. Each resulting question-level metric is then averaged over the questions in its benchmark.

\paragraph{Models.}
Table~\ref{tab:models} lists the reference models and RLVR checkpoints used in the entropy analysis, spanning $1.5$B to $32$B parameters. The reference groups make it possible to compare changes after RLVR at each scale and to compare AceReason-7B and Skywork-OR1-7B against their shared DeepSeek-7B reference.
\begin{table*}[t]
\centering
\caption{Models evaluated in the entropy analysis.}
\label{tab:models}
\resizebox{\textwidth}{!}{
\begin{tabular}{llcc}
\toprule
\textbf{Name} & \textbf{Full Model Name} & \textbf{Type} & \textbf{Parameters} \\
\midrule
DeepSeek-1.5B & \texttt{DeepSeek-R1-Distill-Qwen-1.5B} & Base & 1.5B \\
ProRL-1.5B & \texttt{Nemotron-Research-Reasoning-Qwen-1.5B} & RLVR & 1.5B \\
\midrule
DeepSeek-7B & \texttt{DeepSeek-R1-Distill-Qwen-7B} & Base & 7B \\
AceReason-7B & \texttt{AceReason-Nemotron-7B} & RLVR & 7B \\
Skywork-OR1-7B & \texttt{Skywork-OR1-7B} & RLVR & 7B \\
\midrule
DeepSeek-14B & \texttt{DeepSeek-R1-Distill-Qwen-14B} & Base & 14B \\
AceReason-14B & \texttt{AceReason-Nemotron-14B} & RLVR & 14B \\
\midrule
Qwen2.5-32B & \texttt{Qwen2.5-32B} & Base & 32B \\
DAPO-32B & \texttt{DAPO-Qwen-32B} & RLVR & 32B \\
\bottomrule
\end{tabular}}
\end{table*}

\paragraph{Entropy Computation.}
For token-level entropy, we score each generated completion under its generating model using teacher forcing. A forward pass provides the probability distribution over the vocabulary at each token position, from which we compute the token entropies and their per-completion average. For answer-level entropy, we extract the final answer from each completion using Lighteval~\citep{lighteval} and compute entropy from the empirical frequencies of the extracted answers across the 32 completions.

% \section{Additional Results}
% \begin{figure}[ht]
%     \centering
%     \includegraphics[width=0.45\textwidth]{figures/reasoning_gym/arc_1d_hard.pdf}
%     % \hfill
%     \includegraphics[width=0.45\textwidth]{figures/reasoning_gym/palindrome_generation_hard.pdf}
%     % \hfill
%     \caption{Tasks illustrating {empirical support preservation} and \textcolor{red}{empirical support shrinkage}.}
%     \label{fig:soft_support_more}
% \end{figure}

\section{Reward Reweighting and Sampling Resolution}
\label{sec:theory_rlvr}

\subsection{Effect of Binary-Reward Reweighting}
\label{sec:kl_projection}

Binary-reward reweighting increases access to correct responses by shifting probability toward them. For a fixed question $x$, consider a finite output space $\mathcal{Y}$, a reference distribution $q$, and reward $R(x,y)\in\{0,1\}$. Let $\mathcal{C}_x=\{y:R(x,y)=1\}$ and assume $c=q(\mathcal{C}_x\mid x)\in(0,1)$. We optimize over distributions $\pi\ll q$, meaning that $\pi$ assigns zero probability wherever $q$ does, with $q$ held fixed.

\begin{proposition}[KL projection onto reward-consistent distributions]
\label{thm:kl}
For a target reward $\rho\in[0,1]$, consider
\[
\min_{\pi\ll q}\mathrm{KL}(\pi\|q)
\quad\text{subject to}\quad
\mathbb{E}_{\pi}[R(x,y)]\ge\rho.
\]
If $\rho\le c$, the unique optimizer is $q$. If $c<\rho<1$, the unique optimizer is
\[
\begin{aligned}
\pi_{\beta}(y\mid x)
&=\frac{q(y\mid x)e^{\beta R(x,y)}}{Z_{\beta}(x)},\\
Z_{\beta}(x)&=1-c+e^{\beta}c,
\end{aligned}
\]
where
\[
\beta=\log\frac{\rho(1-c)}{c(1-\rho)}>0.
\]
For $\rho=1$, the optimizer is $q$ conditioned on $\mathcal{C}_x$, obtained as $\beta\to\infty$.
\end{proposition}

\begin{proof}
When $\rho\le c$, $q$ is feasible and uniquely attains zero KL divergence. For $c<\rho<1$, the stated $\beta$ gives $\pi_{\beta}(\mathcal{C}_x\mid x)=\rho$. For every feasible $\pi\ll q$,
\begin{align*}
\mathrm{KL}(\pi\|q)
&=\mathrm{KL}(\pi\|\pi_{\beta})
  +\beta\mathbb{E}_{\pi}[R]-\log Z_{\beta}\\
&\ge\beta\rho-\log Z_{\beta}
 =\mathrm{KL}(\pi_{\beta}\|q).
\end{align*}
Equality requires $\pi=\pi_{\beta}$. For $\rho=1$, feasibility restricts $\pi$ to $\mathcal{C}_x$, where
\[
\mathrm{KL}(\pi\|q)
=\mathrm{KL}(\pi\|q(\cdot\mid x,\mathcal{C}_x))-\log c.
\]
The unique minimizer is therefore the conditional reference distribution.
\end{proof}

For any finite $\beta>0$, the same exponential tilt uniquely maximizes the KL-regularized objective
\[
\mathbb{E}_{\pi}[R(x,y)]-\beta^{-1}\mathrm{KL}(\pi\|q),
\]
which equals $\beta^{-1}(\log Z_{\beta}-\mathrm{KL}(\pi\|\pi_{\beta}))$. For $c<\rho<1$, the constrained formulation selects $\beta$ to attain the target reward $\rho$; the regularized formulation sets $\beta$ directly. Larger $\beta$ gives greater weight to correct responses. Exponential-tilting policy improvement also appears in KL-regularized contextual bandits and RLHF~\citep{zhao2024sharp}.

\begin{corollary}[Correctness and the zero-penalty limit]
\label{cor:kl_free_short}
For the fixed-reference binary-reward tilt and finite $\beta\ge0$,
\[
\pi_{\beta}(\mathcal{C}_x\mid x)
=\frac{e^{\beta}c}{1-c+e^{\beta}c}\ge c.
\]
Every correct completion is multiplied by the same factor $e^{\beta}/(1-c+e^{\beta}c)\ge1$. Exact support and the conditional distribution among correct completions remain unchanged. As $\beta\to\infty$,
\[
\pi_{\beta}(y\mid x)\longrightarrow
\frac{q(y\mid x)\mathbf{1}\{y\in\mathcal{C}_x\}}{c}.
\]
For independent samples from these distributions, expected \texttt{pass@k} cannot decrease for any $k\ge1$.
\end{corollary}

\begin{proof}
The expression for $Z_{\beta}$ gives the total correct mass and the common multiplier directly. The multiplier is at least one because $e^{\beta}\ge1$. Conditioning on correctness cancels it, and taking $\beta\to\infty$ gives the stated limit. If the per-sample success probability is $s$, the probability of at least one success in $k$ independent samples is $1-(1-s)^k$, which is nondecreasing in $s$.
\end{proof}

Applied independently across questions with direct sampling from $q$ and $\pi_{\beta}$, this update gives $A_{\epsilon}(q)\subseteq A_{\epsilon}(\pi_{\beta})$ at every positive threshold. A question enters the accessible set when its reweighted correct mass crosses $\epsilon$. Thus, reward reweighting can expand practical access to solutions already present in the reference distribution's tail.

\paragraph{Entropy under reweighting.}
The same reward tilt can broaden or concentrate the output distribution, depending on the initial mass assigned to correct responses. For $q=(0.9,0.1)$, $R=(0,1)$, and $\beta=\log9$, the updated distribution is $\pi_{\beta}=(0.5,0.5)$. Correctness probability rises from $0.1$ to $0.5$, while entropy rises from approximately $0.325$ to $0.693$ nats. Starting instead from $q=(0.5,0.5)$ with the same reward and $\beta$ gives $\pi_{\beta}=(0.1,0.9)$: correctness probability rises to $0.9$ and entropy falls to $0.325$ nats.

\subsection{Finite-Budget Resolution of Verified Accessibility}
\label{appendix:threshold}

For a fixed question, verifier, and decoding policy, let $s$ be the probability that one sampled response passes verification. With $n$ independent samples, the probability of at least one verified success is $1-(1-s)^n$. After zero successes, an exact one-sided $(1-\alpha)$ binomial upper confidence bound is
\[
u_n=1-\alpha^{1/n}\le\frac{-\log\alpha}{n}.
\]
The exact bound solves $(1-u_n)^n=\alpha$; the inequality follows from $1-e^{-a}\le a$ for $a\ge0$.

\paragraph{Example.} For a question with zero verified successes in $8192$ independent samples, the one-sided $95\%$ upper confidence bound on its success probability is $u_n\approx3.656\times10^{-4}$. The bound decreases approximately as $1/n$, quantifying how larger sampling budgets resolve rarer verified successes.

\section{Numerical Values for Reasoning Gym Curves}
\label{app:reasoning_gym_values}
Tables~\ref{tab:curve-values-preservation-a}--\ref{tab:curve-values-shrinkage} report the Base and ProRL pass@$k$ values in Figures~\ref{fig:soft_support_in}--\ref{fig:soft_support_shrinkage}. A dash marks a sampling budget not shown in the corresponding panel.

\begin{table*}[t]
\centering
\small
\caption{Preservation, top row: pass@$k$ values for Figure~\ref{fig:soft_support_in}. Each pair of columns compares the Base model with ProRL at the same sampling budget $k$. A dash marks a sampling budget not shown in the corresponding panel.}
\label{tab:curve-values-preservation-a}
\setlength{\tabcolsep}{3pt}
\renewcommand{\arraystretch}{1.05}
\begin{tabular*}{\textwidth}{@{\extracolsep{\fill}}r rr rr rr@{}}
\toprule
$k$ & \multicolumn{2}{c}{\texttt{dice}} & \multicolumn{2}{c}{\texttt{prime\_factorization\_hard}} & \multicolumn{2}{c}{\texttt{palindrome\_generation}} \\
\cmidrule(lr){2-3}\cmidrule(lr){4-5}\cmidrule(lr){6-7}
 & Base & ProRL & Base & ProRL & Base & ProRL \\
\midrule
1 & 0.000 & 0.180 & 0.832 & 1.000 & 0.154 & 0.923 \\
2 & 0.000 & 0.460 & 0.941 & 1.000 & 0.235 & 0.980 \\
4 & 0.000 & 0.820 & 0.970 & 1.000 & 0.413 & 0.980 \\
8 & 0.020 & 0.920 & 0.980 & 1.000 & 0.688 & 0.980 \\
16 & 0.060 & 0.980 & 0.980 & 1.000 & 0.825 & 0.980 \\
32 & 0.100 & 1.000 & 1.000 & 1.000 & 0.884 & 1.000 \\
64 & 0.200 & 1.000 & 1.000 & 1.000 & 0.981 & 1.000 \\
128 & 0.320 & 1.000 & 1.000 & 1.000 & 1.000 & 1.000 \\
256 & 0.360 & 1.000 & 1.000 & 1.000 & 1.000 & 1.000 \\
512 & 0.460 & 1.000 & 1.000 & 1.000 & 1.000 & 1.000 \\
1024 & 0.480 & 1.000 & 1.000 & 1.000 & 1.000 & 1.000 \\
2048 & 0.540 & 1.000 & \textemdash & \textemdash & \textemdash & \textemdash \\
4096 & 0.860 & 1.000 & \textemdash & \textemdash & \textemdash & \textemdash \\
8192 & 1.000 & 1.000 & \textemdash & \textemdash & \textemdash & \textemdash \\
\bottomrule
\end{tabular*}
\end{table*}

\begin{table*}[t]
\centering
\small
\caption{Preservation, middle row: pass@$k$ values for Figure~\ref{fig:soft_support_in}. Each pair of columns compares the Base model with ProRL at the same sampling budget $k$. A dash marks a sampling budget not shown in the corresponding panel.}
\label{tab:curve-values-preservation-b}
\setlength{\tabcolsep}{3pt}
\renewcommand{\arraystretch}{1.05}
\begin{tabular*}{\textwidth}{@{\extracolsep{\fill}}r rr rr rr@{}}
\toprule
$k$ & \multicolumn{2}{c}{\texttt{advanced\_geometry}} & \multicolumn{2}{c}{\texttt{advanced\_geometry\_hard}} & \multicolumn{2}{c}{\texttt{prime\_factorization}} \\
\cmidrule(lr){2-3}\cmidrule(lr){4-5}\cmidrule(lr){6-7}
 & Base & ProRL & Base & ProRL & Base & ProRL \\
\midrule
1 & 0.611 & 0.865 & 0.157 & 0.551 & 0.980 & 1.000 \\
2 & 0.690 & 0.904 & 0.275 & 0.728 & 0.980 & 1.000 \\
4 & 0.828 & 0.923 & 0.414 & 0.845 & 0.980 & 1.000 \\
8 & 0.886 & 1.000 & 0.511 & 0.922 & 1.000 & 1.000 \\
16 & 0.905 & 1.000 & 0.686 & 0.961 & 1.000 & 1.000 \\
32 & 0.924 & 1.000 & 0.705 & 0.980 & 1.000 & 1.000 \\
64 & 0.943 & 1.000 & 0.804 & 0.980 & 1.000 & 1.000 \\
128 & 0.943 & 1.000 & 0.804 & 0.980 & 1.000 & 1.000 \\
256 & 0.962 & 1.000 & 0.823 & 0.980 & 1.000 & 1.000 \\
512 & 0.981 & 1.000 & 0.842 & 0.980 & 1.000 & 1.000 \\
1024 & 0.981 & 1.000 & 0.842 & 0.980 & 1.000 & 1.000 \\
2048 & 1.000 & 1.000 & 0.941 & 1.000 & \textemdash & \textemdash \\
4096 & \textemdash & \textemdash & 1.000 & 1.000 & \textemdash & \textemdash \\
\bottomrule
\end{tabular*}
\end{table*}

\begin{table*}[t]
\centering
\small
\caption{Preservation, bottom row: pass@$k$ values for Figure~\ref{fig:soft_support_in}. Each pair of columns compares the Base model with ProRL at the same sampling budget $k$. A dash marks a sampling budget not shown in the corresponding panel.}
\label{tab:curve-values-preservation-c}
\setlength{\tabcolsep}{3pt}
\renewcommand{\arraystretch}{1.05}
\begin{tabular*}{\textwidth}{@{\extracolsep{\fill}}r rr rr rr@{}}
\toprule
$k$ & \multicolumn{2}{c}{\texttt{graph\_color}} & \multicolumn{2}{c}{\texttt{graph\_color\_hard}} & \multicolumn{2}{c}{\texttt{graph\_color\_vertex15}} \\
\cmidrule(lr){2-3}\cmidrule(lr){4-5}\cmidrule(lr){6-7}
 & Base & ProRL & Base & ProRL & Base & ProRL \\
\midrule
1 & 0.140 & 1.000 & 0.120 & 0.960 & 0.100 & 1.000 \\
2 & 0.241 & 1.000 & 0.121 & 1.000 & 0.161 & 1.000 \\
4 & 0.421 & 1.000 & 0.221 & 1.000 & 0.222 & 1.000 \\
8 & 0.742 & 1.000 & 0.462 & 1.000 & 0.382 & 1.000 \\
16 & 0.960 & 1.000 & 0.582 & 1.000 & 0.602 & 1.000 \\
32 & 0.980 & 1.000 & 0.802 & 1.000 & 0.742 & 1.000 \\
64 & 1.000 & 1.000 & 0.822 & 1.000 & 0.841 & 1.000 \\
128 & 1.000 & 1.000 & 0.881 & 1.000 & 0.861 & 1.000 \\
256 & 1.000 & 1.000 & 0.901 & 1.000 & 0.941 & 1.000 \\
512 & 1.000 & 1.000 & 0.941 & 1.000 & 1.000 & 1.000 \\
1024 & 1.000 & 1.000 & 0.980 & 1.000 & 1.000 & 1.000 \\
2048 & \textemdash & \textemdash & 1.000 & 1.000 & \textemdash & \textemdash \\
\bottomrule
\end{tabular*}
\end{table*}

\begin{table*}[t]
\centering
\small
\caption{Expansion: pass@$k$ values for Figure~\ref{fig:soft_support_out}. Each pair of columns compares the Base model with ProRL at the same sampling budget $k$. A dash marks a sampling budget not shown in the corresponding panel.}
\label{tab:curve-values-expansion}
\setlength{\tabcolsep}{3pt}
\renewcommand{\arraystretch}{1.05}
\begin{tabular*}{\textwidth}{@{\extracolsep{\fill}}r rr rr rr@{}}
\toprule
$k$ & \multicolumn{2}{c}{\texttt{arc\_1d}} & \multicolumn{2}{c}{\texttt{boxnet}} & \multicolumn{2}{c}{\texttt{graph\_color\_vertex20}} \\
\cmidrule(lr){2-3}\cmidrule(lr){4-5}\cmidrule(lr){6-7}
 & Base & ProRL & Base & ProRL & Base & ProRL \\
\midrule
1 & 0.010 & 0.040 & 0.024 & 0.097 & 0.000 & 0.881 \\
2 & 0.037 & 0.109 & 0.048 & 0.151 & 0.000 & 0.980 \\
4 & 0.064 & 0.158 & 0.063 & 0.179 & 0.001 & 0.980 \\
8 & 0.097 & 0.297 & 0.083 & 0.210 & 0.023 & 1.000 \\
16 & 0.128 & 0.375 & 0.102 & 0.239 & 0.044 & 1.000 \\
32 & 0.197 & 0.499 & 0.131 & 0.258 & 0.126 & 1.000 \\
64 & 0.295 & 0.526 & 0.148 & 0.283 & 0.187 & 1.000 \\
128 & 0.348 & 0.560 & 0.185 & 0.294 & 0.247 & 1.000 \\
256 & 0.397 & 0.571 & 0.209 & 0.325 & 0.287 & 1.000 \\
512 & 0.447 & 0.577 & 0.216 & 0.343 & 0.366 & 1.000 \\
1024 & 0.460 & 0.578 & 0.253 & 0.366 & 0.406 & 1.000 \\
2048 & 0.570 & 0.680 & 0.289 & 0.437 & 0.762 & 1.000 \\
4096 & \textemdash & \textemdash & 0.308 & 0.443 & 0.901 & 1.000 \\
8192 & \textemdash & \textemdash & 0.332 & 0.475 & \textemdash & \textemdash \\
16384 & \textemdash & \textemdash & 0.345 & 0.492 & \textemdash & \textemdash \\
\bottomrule
\end{tabular*}
\end{table*}

\begin{table*}[t]
\centering
\small
\caption{Shrinkage: pass@$k$ values for Figure~\ref{fig:soft_support_shrinkage}. Each pair of columns compares the Base model with ProRL at the same sampling budget $k$. A dash marks a sampling budget not shown in the corresponding panel.}
\label{tab:curve-values-shrinkage}
\setlength{\tabcolsep}{3pt}
\renewcommand{\arraystretch}{1.05}
\begin{tabular*}{\textwidth}{@{\extracolsep{\fill}}r rr rr rr@{}}
\toprule
$k$ & \multicolumn{2}{c}{\texttt{leg\_counting}} & \multicolumn{2}{c}{\texttt{family\_relationships}} & \multicolumn{2}{c}{\texttt{power\_function}} \\
\cmidrule(lr){2-3}\cmidrule(lr){4-5}\cmidrule(lr){6-7}
 & Base & ProRL & Base & ProRL & Base & ProRL \\
\midrule
1 & 0.040 & 0.280 & 0.140 & 0.840 & 0.465 & 0.624 \\
2 & 0.040 & 0.300 & 0.220 & 0.860 & 0.604 & 0.802 \\
4 & 0.120 & 0.300 & 0.380 & 0.860 & 0.723 & 0.842 \\
8 & 0.140 & 0.300 & 0.560 & 0.860 & 0.842 & 0.901 \\
16 & 0.160 & 0.300 & 0.680 & 0.860 & 0.941 & 0.901 \\
32 & 0.241 & 0.300 & 0.780 & 0.860 & 1.000 & 0.901 \\
64 & 0.280 & 0.300 & 0.860 & 0.860 & 1.000 & 0.901 \\
128 & 0.300 & 0.300 & 0.900 & 0.860 & 1.000 & 0.901 \\
256 & 0.320 & 0.300 & 0.900 & 0.860 & 1.000 & 0.901 \\
512 & 0.340 & 0.300 & 0.960 & 0.860 & 1.000 & 0.901 \\
1024 & 0.400 & 0.300 & 1.000 & 0.880 & 1.000 & 0.901 \\
\bottomrule
\end{tabular*}
\end{table*}

\end{document}